\documentclass[10pt,twocolumn,letterpaper]{article}

\usepackage{cvpr}              

\definecolor{myred}{RGB}{184, 84, 80}
\definecolor{mygreen}{RGB}{18.4, 89.9, 63.5}

\usepackage{siunitx}
\usepackage{multirow}

\DeclareSIUnit{\fps}{fps}
\DeclareSIUnit{\px}{px}
\DeclareSIUnit{\inch}{in} 

\definecolor{cvprblue}{rgb}{0.21,0.49,0.74}
\usepackage[pagebackref,breaklinks,colorlinks,allcolors=cvprblue]{hyperref}

\def\confName{CVPR}
\def\confYear{2026}

\title{Beyond Scanpaths: Graph-Based Gaze Simulation in Dynamic Scenes}

\newlength{\authurlgap}
\newcommand{\urlfontsize}{11}
\newcommand{\urlbaselineskip}{9}
\newlength{\urlgap}
\author{
Luke Palmer$^{1}$\thanks{These authors contributed equally to this work} \quad
Petar Palasek$^{1}$\footnotemark[1] \quad
Hazem Abdelkawy$^2$ \\
$^1$GlimpseML \quad $^2$Toyota Motor Europe \\
{\tt\small \{luke, petar\}@glimpse.ml, hazem.abdelkawy@toyota-europe.com} \\[\authurlgap]
{\fontsize{\urlfontsize}{\urlbaselineskip}\selectfont
\href{https://glimpse.ml/beyond-scanpaths}{\texttt{glimpse.ml/beyond-scanpaths}}
}
\vspace{\urlgap}
}

\begin{document}

\maketitle
\begin{abstract}
Accurately modelling human attention is essential for numerous computer vision applications, particularly in the domain of automotive safety. Existing methods typically collapse gaze into saliency maps or scanpaths, treating gaze dynamics only implicitly. We instead formulate gaze modelling as an autoregressive dynamical system and explicitly unroll raw gaze trajectories over time, conditioned on both gaze history and the evolving environment.
Driving scenes are represented as gaze-centric graphs processed by the Affinity Relation Transformer (ART), a heterogeneous graph transformer that models interactions between driver gaze, traffic objects, and road structure. We further introduce the Object Density Network (ODN) to predict next-step gaze distributions, capturing the stochastic and object-centric nature of attentional shifts in complex environments. We also release Focus100, a new dataset of raw gaze data from 30 participants viewing egocentric driving footage. Trained directly on raw gaze, without fixation filtering, our unified approach produces more natural gaze trajectories, scanpath dynamics, and saliency maps than existing attention models, offering valuable insights for the temporal modelling of human attention in dynamic environments. 

\end{abstract}    
\section{Introduction}
\label{sec:intro}

\begin{figure}[htb]
\centering
\includegraphics[width=1\linewidth]{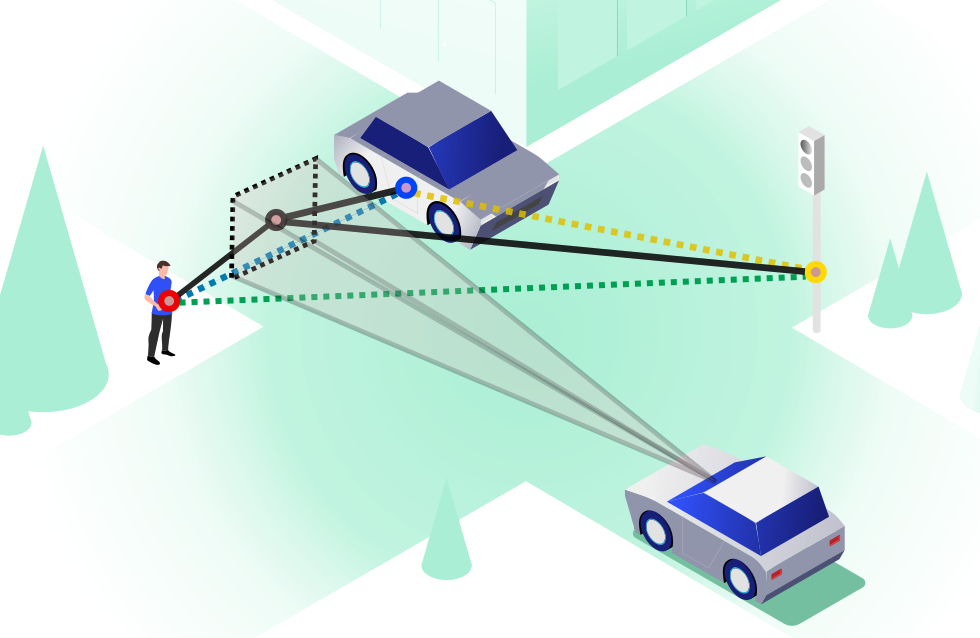}
\caption{To model driver attention as part of a dynamical system we encode traffic scenes as heterogeneous scene graphs with nodes corresponding to road structure, driving-relevant objects, and \textit{egocentric gaze}, representing the driver's foveated field-of-view. The Affinity Relation Transformer processes these graphs to predict next-step gaze probability distributions. Our dynamical systems approach generates state of the art \textcolor{myred}{gaze timeseries}, \textcolor{mygreen}{scanpaths}, and \textcolor{cvprblue}{saliency maps} from a \textit{single model}.}
\label{fig:second_column_image}
\end{figure}

Understanding and predicting human attention allocation in dynamic environments underpins applications such as image and video compression \cite{hadizadeh2013saliency}, realistic avatar animation \cite{lee2002eyes, admoni2017social}, and foveated rendering \cite{patney2016towards, chao2021transformer}. In the driving domain, gaze modelling informs assessments of driver situational awareness \cite{fletcher2009driver, hofbauer2020measuring, rezaei2014look, ma2019gazefcw}, a critical factor in automotive safety \cite{stanton2009human}.  Prior works simplify gaze into  saliency maps \cite{palazzi2018predicting, fang2021dada, xia2019predicting} or discrete fixation sequences \cite{huang2024driver, yang2020predicting}, obscuring natural dynamics (\eg smooth pursuit \cite{robinson1965mechanics, buizza2021computer}). Moreover, the necessary fixation filtering introduces artefacts and data loss for video stimuli where fixation algorithms are unreliable \cite{andersson2017one}. We instead learn the underlying gaze-generating process from raw sequences, \textit{without fixation filtering}, enabling a single model trained once to output raw trajectories and, via training-free post-processing, scanpaths and saliency maps, achieving state-of-the-art performance across all three representations.

Towards this unified approach, we formulate gaze prediction as graph-based simulation (GBS; \cite{sanchez2020learning}), modelling its spatio-temporal evolution as an active agent within the visual environment. GBS captures complex relationships in structured data by representing physical systems as graphs with objects/agents as nodes and physical relations as edges. Often leveraging graph neural networks (GNNs), it has been successful in simulating dynamic particle systems (\eg water, sand, cloth \cite{Luo2024care, pfaff2020learning, sanchez2020learning, xu2024learning}) and motion trajectories \cite{Shi2021sgcn, lian2023ptp, mo2021heterogeneous}). Recent works use graph transformers as the GBS backbone \cite{Shao2022transformer, janny2023eagle, jia2023hdgt}. Our work extends GBS to gaze simulation, marking the first application in video and driving settings, contrasting with prior gaze generation work in static image free-viewing \cite{jiao2024diffgaze}.

We propose a gaze-centric spatiotemporal heterogeneous graph representation of driving scenes for our GBS approach. Task-relevant elements (\eg cars, pedestrians, signage, road layout) are represented as nodes in a heterogeneous scene graph, connected spatially and temporally across frames. Nodes contain feature vectors based on position, time, and appearance, while edges represent relative position and similarity. Additionally, a \textit{gaze node} at each timestep represents the driver's foveated field of view (see Figure \ref{fig:second_column_image}). By linking gaze nodes over time, we predict future gaze autoregressively, conditioning on both gaze and environment history, thereby modelling temporal interactions between the driver’s gaze and traffic objects. To process this graph, we introduce a novel heterogeneous graph transformer, the \textit{Affinity Relation Transformer} (ART), which encodes relative features from graph edges and incorporates them directly into the graph attention mechanism.

The ART module's output, which captures interactions within the gaze-centric graph, feeds into our \textit{Object Density Network} (ODN) for gaze prediction. Unlike existing pixel-level predictions, we take an object-centric approach, motivated by findings that human attention in complex tasks like driving is guided by object relevance \cite{roth2023objects, nuthmann2010object, borji2013look}. ODN models the probability of gaze focusing on each scene graph node, forming a Gaussian mixture where each node contributes a component reflecting its salience. This differs from typical MDN approaches for saliency and scanpath prediction \cite{De2022scanpathnet, sun2019visual, quan2024pathformer3d}, which rely on a fixed number of Gaussian components; instead, ODN adapts the mixture components based on the scene's content and complexity. The ODN also provides an interpretable fixation mechanism: high mixing weight on the gaze node keeps gaze near its current location, while high weight on environment nodes shifts gaze toward relevant traffic entities or regions.

Finally, recognising the limitations of several existing datasets (\eg \cite{fang2021dada, xia2019predicting}), which provide only aggregated saliency maps as ground truth, we introduce Focus100, an in-lab gaze dataset collected across 30 subjects. Focus100 provides raw gaze across various driving scenarios, enabling precise evaluation of our method. Our method produces more natural gaze sequences, scanpaths and saliency maps on Focus100 compared to existing approaches.

In summary, our contributions are fourfold: (1) a gaze-centric, dynamic, spatiotemporal heterogeneous graph representation for driving scenes; (2) a heterogeneous graph transformer module with relative affinity encoding; (3) an object-centric mixture density task head to model stochastic human attention shifts; and (4) a new driver gaze dataset, Focus100, to validate our approach and spur further research in this critical area.

\section{Related work}
\label{sec:related}

We organise related work into three areas: saliency and scanpath prediction, attention modelling in driving, and graph-based simulation.

\paragraph{Saliency and Scanpaths.}

While attention modelling has focused primarily on static images, using heuristic \cite{itti1998model, bruce2005saliency, le2006coherent, garcia2009decorrelation, harel2006graph, kootstra2008paying, murray2011saliency, seo2009static} and deep learning approaches \cite{cornia2016deep, kruthiventi2017deepfix, kummerer2014deep, pan2017salgan, pan2016shallow, linardos2021deepgaze, djilali2024learning, kummerer2022deepgaze}, our work is concerned with the more complex task of predicting gaze sequences in dynamic, task-driven video settings. In video-based saliency prediction, models have evolved to capture temporal dynamics using CNN-LSTM, vision transformers, and adversarial models \cite{ li2013learning, li2021eye, tavakoli2019digging, tsiami2020stavis, wang2021spatio, zhang2017deep}. However, these methods produce aggregated gaze probability maps, overlooking the temporal dynamics of gaze.

Scanpath prediction, which models sequences of fixations, is less explored, particularly in video. Many approaches use diffusion models, transformers, Markov models, and reinforcement learning to generate fixation sequences \cite{mondal2023gazeformer, chen2021predicting, qiu2023simulating, yang2020predicting, sui2023scandmm}, while several incorporate Gaussian mixtures as a probabilistic model for fixation generation \cite{sun2019visual, De2022scanpathnet, quan2024pathformer3d}. Video scanpath modelling has also been approached in VR and panoramic video settings \cite{fan2017fixation, xu2021spherical, rondon2021track, li2019very, fan2024learned}, combining fixation history with image features for multimodal inputs. While \cite{jiao2024diffgaze} applied diffusion models for continuous gaze generation in image free-viewing and \cite{xu2018gaze} used LSTMs to generate gaze over 360$^{\circ}$ imagery, our work is the first to tackle continuous gaze sequence generation in task-driven video, particularly within a driving context.

\paragraph{Attention in Driving.}

Several methods have been developed to model the spatial distribution of drivers' attention in traffic scenes, producing 2D saliency maps from sequences of images. Methods in this area have employed optical flow \cite{palazzi2018predicting, tavakoli2019digging, ning2019efficient, gopinath2021maad}, scene dynamics \cite{amadori2021hammerdrive, baee2021medirl}, traditional saliency algorithms \cite{deng2016does, deng2017learning, borji2011computational, borji2012probabilistic, borji2013look}, 3D convolutional networks \cite{palmer2017predicting, zhang2020interaction, palazzi2017learning, xia2019predicting, tawari2018learning}, and graph convolutional networks (GCNs) \cite{fang2021dada} to predict gaze distributions in driving. Conditioning on driver intention has recently been explored \cite{kotseruba2024scout+}; although orthogonal to this study, it is a natural avenue for future work. In contrast, the prediction of scanpaths in driving has received limited attention, with only one identified work \cite{huang2024driver} that predicts spatial fixation sequences (without durations) using a CNN-Transformer architecture trained via inverse reinforcement learning.

Several driver attention datasets are limited by simulator settings \cite{borji2011computational, taamneh2017multimodal} or release only aggregated attention maps \cite{fang2021dada, xia2019predicting, deng2019drivers, deng2016does}, restricting analyses to the spatial characteristics of attention allocation. The DR(eye)VE dataset \cite{palazzi2018predicting}, despite releasing raw gaze sequences, is hindered by temporal misalignment and limited scenario complexity and diversity \cite{kotseruba2024data}. MAAD \cite{gopinath2021maad} corrects these issues through in-lab gaze tracking but only for a small subset of DR(eye)VE data. \textit{Focus100} spans a broader range of traffic scenarios, includes hazard annotations, and provides more than twice the driving footage and three times the gaze data of MAAD.

\paragraph{Graph Representation and Simulation.}
Graph-based simulation (GBS) has been used to model complex dynamics in physical systems such as fluids, cloth, and sand, due to its ability to capture interactions between entities in a flexible and structured manner \cite{Luo2024care, pfaff2020learning, sanchez2020learning, xu2024learning}. Recently, GBS has been shown capable of modelling discontinuous and stochastic dynamics \cite{allen2023graph, sun2024unifying}. The dynamics of human gaze, characterised by discontinuous and stochastic shifts \cite{Brockmann2000ecology, boccignone2013ecological} fit well within this approach, and ours is the first work to apply GBS to model human attention allocation.

In computer vision, graphs have been leveraged in several domains to represent scenes and predict object dynamics \cite{Johnson2015image, wang2018videos, mottaghi2016happens, wang2021joint, wang2021object}, while recent works have utilised GBS to predict future trajectories of traffic objects \cite{liu2022social, yu2020spatio, jia2023hdgt, wang2023spatio, zhang2022trajectory, shi2023trajectory, mo2021heterogeneous}. The work in \cite{jia2023hdgt}, for instance, constructs heterogeneous traffic graphs and applies graph transformers for future trajectory prediction. Our approach is the first to integrate driver attention into this framework, introducing a dedicated attention node to model interactions between gaze and spatio-temporal scene graphs. Inspired by relative positional encoding in language and image models \cite{shaw2018self, wu2021rethinking, zhang2024real} we also propose the Affinity Relation Transformer (ART) to inject arbitrary relational information directly into a heterogeneous graph transformer module.

\begin{figure*}[ht]
\centering

\includegraphics[width=\linewidth]{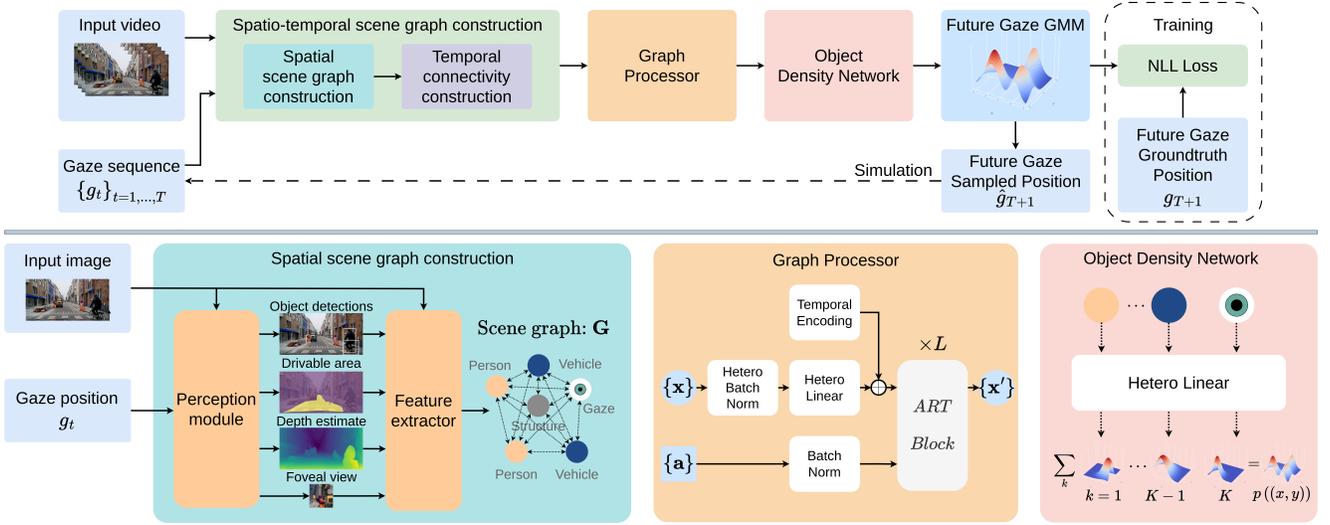}

\caption{Using upstream perception modules and observed gaze position, synchronised driving video and gaze are converted into a spatiotemporal heterogeneous scene graph with nodes for traffic agents, road structure, and driver foveal view. Each node is assigned a feature vector including appearance and depth, while edges represent the spatiotemporal differences and appearance similarities between nodes. Scene graphs are processed by Affinity Relation Transformer (ART) blocks before an Object Density Network (ODN) predicts a Gaussian-mixture distribution for the next gaze position. We train the model with negative log-likelihood of ground-truth gaze under the predicted mixture. For simulation we employ autoregressive rollout by sampling from the mixture, updating the graph with the sampled position, and repeating; simulated gaze sequences can then be post-processed into scanpaths and saliency maps without additional training.}
\label{fig:pipeline}
\end{figure*}

\section{Method}

We model the driver's future gaze using a dynamic, heterogeneous, spatio-temporal graph representation of the driving scene, incorporating historical gaze positions as special nodes at each input timestep (\cref{fig:pipeline}). 
The initial graph is processed by a \textit{Graph Processor} (\textit{GP}), formed of a stack of \textit{Affinity Relation Transformers} (\textit{ART}), and an \textit{Object-based mixture Density Network (ODN)}, which outputs a Gaussian mixture model (\textit{GMM}) for the next-step gaze position.

\subsection{Spatio-Temporal Heterogeneous Scene Graph}
\label{ref:scene_graph}
We wish to represent a sequence of $T$ input frames, $\mathcal{I}=[I_1, ..., I_{T}]$, by a spatio-temporal heterogeneous scene graph $\mathbf{G}$. 
We follow \cite{hu2020heterogeneous} and define a heterogeneous graph as a directed graph $\mathcal{G}=\left(\mathcal{V}, \mathcal{E}, \mathcal{A}, \mathcal{R}\right)$, where $\mathcal{V}$ denotes the set of nodes, $\mathcal{E}$ denotes the set of edges, each node $\mathbf{v} \in \mathcal{V}$ is assigned a node type $\tau(\mathbf{v}): \mathcal{V} \mapsto \mathcal{A}$, and each edge $\mathbf{e} \in \mathcal{E}$ is assigned an edge type $\phi(\mathbf{e}) \mapsto \mathcal{R}$.
We use $\mathbf{e}_{\mathbf{v}_j \rightarrow \mathbf{v}_i}$ to refer to an edge from node $\mathbf{v}_j$ to node $\mathbf{v}_i$.

\paragraph{Nodes.} Each frame $I_t$ of the input sequence $\mathcal{I}$ contains a varying number of traffic-related entities such as cars, pedestrians, and traffic lights. We build a spatio-temporal heterogeneous scene graph $\mathbf{G}$ by representing each entity at timestep $t\in[1, T]$ as a node. Each node's feature vector $\mathbf{x}$ includes the 2D position (bounding box centre), bounding box shape, detection score, appearance vector, depth estimate and one-hot label encoding. We additionally include a special \textbf{gaze node} at each timestep, representing the driver's foveal view; it uses the same feature definition, with its bounding box centred at the measured gaze location and set to a fixed fraction of the input shape (20\% height, 10\% width), and its appearance extracted from the corresponding image crop. A \textit{structure} node encoding the drivable area in frame $I_t$ is also introduced at each timestep. Finally, heterogeneous node types are assigned by grouping detector labels into dynamically meaningful categories following \cite{jia2023hdgt} so that the model can allocate type-specific parameters to capture each category’s dynamics (see \cref{tab:node_types}).

\begin{table}[ht]
\footnotesize
\centering
\caption{Node type descriptions.}
\begin{tabular}{cl}
\toprule
\textbf{Node type}    & \textbf{Description}                                                                                                                                          \\ \midrule
\textit{vehicle}      & \begin{tabular}[c]{@{}l@{}}Detections with labels `\textit{car}', `\textit{bicycle}', \\  `\textit{motorcycle}', `\textit{bus}', `\textit{truck}'.\end{tabular} \\ 
\textit{person}       & Detections with label `\textit{person}'.                                                                                                                      \\ 
\textit{static}       & Detections with labels `\textit{traffic light}', `\textit{stop sign}'.                                                                                        \\ 
\textit{gaze}         & Special node representing the gaze state.          \\ 

\textit{structure} & Special node representing the road structure.                                                                                                                   \\ \bottomrule
\end{tabular}
\label{tab:node_types}
\end{table}

\paragraph{Edges.} We define two categories of edges: \textit{spatial}, connecting nodes within the same timestep $t$, and \textit{temporal}, connecting nodes across different timesteps. At each timestep, all node pairs $(\mathbf{v}_i, \mathbf{v}_j)$ are connected by two directed spatial edges, $\mathbf{e}_{\mathbf{v}_j \rightarrow \mathbf{v}_i}$ and $\mathbf{e}_{\mathbf{v}_i \rightarrow \mathbf{v}_j}$, modelling interactions in both directions. 
Temporal edges connect nodes in a causal way, only from past to future timesteps.
Nodes are connected temporally if their timestep difference is included in a predefined set $\mathcal{T}_d$.
The type of each edge is defined by a triplet formed of the source node type, edge category, and the destination node type: $\phi(\mathbf{e}_{\mathbf{v}_j \rightarrow \mathbf{v}_i}):=\left(\tau(\mathbf{v}_j), C(\mathbf{e}_{\mathbf{v}_j \rightarrow \mathbf{v}_i}), \tau(\mathbf{v}_i)\right)$, where $C(\mathbf{e}_{\mathbf{v}_j \rightarrow \mathbf{v}_i}) \in \{\texttt{spatial}, \texttt{temporal}\}$.

Each edge is assigned a feature vector $\mathbf{a}_{i,j}$ modelling a generalised \textit{affinity} between nodes $\mathbf{v}_j$ and $\mathbf{v}_i$ across space, time and appearance. This vector includes differences in 3D position, timestep differences, and cosine similarity between destination and source node appearance vectors. Edges between gaze nodes and object nodes allow information flow about objects previously attended to and those attended at timestep $t$. Combining this with the gaze history encoded across gaze nodes gives context to autoregressively predict the next-step gaze distribution.

\subsection{Graph Processor}

\label{ref:art}

\paragraph{Input Embeddings.}
Given the spatio-temporal heterogeneous scene graph $\mathbf{G}$ from \cref{ref:scene_graph}, node features $\mathbf{x}$ and edge features $\mathbf{a}$ are batch-normalised \cite{ioffe2015batch}. The node vectors are then projected with node-type-specific linear layers and augmented with a temporal encoding of alternating sine/cosine waves following \cite{vaswani2017attention}. These embedded features are the inputs to a stack of $L$ Affinity Relation Transformer (ART) blocks.

\paragraph{ART.} We first recall the generic message-passing model and then specialise it to heterogeneous graph transformers (HGT) \cite{hu2020heterogeneous} and our Affinity Relation Transformer (ART) module. A message passing GNN layer \cite{bronstein2021geometric} updates each node vector representation $\mathbf{x}_i$ as:
\begin{equation}
\label{ref:message_passing}
\mathbf{x}'_i = \Phi \left( \mathbf{x}_i,  \bigoplus_{j \in \mathcal{N}_i}a(\mathbf{x}_i, \mathbf{x}_j) \psi (\mathbf{x}_i, \mathbf{x}_j) \right),
\end{equation}
where $\Phi$, $\psi$ and $a$ denote the learnable \textit{update}, \textit{message}, and \textit{attention} operations, and $\bigoplus$ is a permutation invariant \textit{aggregation} operator (\eg \textit{sum}, \textit{mean}, \textit{max}) operating over the neighbourhood of $\mathbf{x}_i$, $\mathcal{N}_i$.

HGT \cite{hu2020heterogeneous} defines type-specific scaled dot-product attention:
\begin{equation}
a(\mathbf{x}_i, \mathbf{x}_j) = \xi_j \left( \frac{\mathbf{Q}_i\mathbf{K}_j^T}{\sqrt{d}} \right),
\end{equation}
and the \textit{query}, \textit{key}, and \textit{value} vectors as:
\begin{align}
\label{eq:qkv}
\mathbf{Q}_i &= \mathbf{x}_i\mathbf{W}_Q^{\tau} + \mathbf{b}_Q^{\tau},  \\
\mathbf{K}_j &= \left(\mathbf{x}_j\mathbf{W}_K^{\tau} + \mathbf{b}_K^{\tau}\right) \mathbf{W}_K^{\phi}, \\
\mathbf{V}_j&=\psi (\mathbf{x}_i, \mathbf{x}_j)=\left(\mathbf{x}_j\mathbf{W}_V^{\tau} + \mathbf{b}_V^{\tau}\right)\mathbf{W}_V^{\phi}.
\end{align}
$\mathbf{W}_Q^{\tau}$ and $\mathbf{W}_K^{\tau}$ are target node-type-specific linear maps, $\mathbf{W}_V^{\tau}$ is a source node-type-specific \textit{value} map, and $\mathbf{W}_K^{\phi}$ and $\mathbf{W}_V^{\phi}$ are edge-type-specific maps depending on $\phi(\mathbf{e}_{\mathbf{v}_j \rightarrow\mathbf{v}_i})$. $\mathbf{b}_Q^{\tau}$, $\mathbf{b}_K^{\tau}$, and $\mathbf{b}_V^{\tau}$ are source node-type-specific biases, $d$ is the dimension of $\mathbf{Q}_i$ and $\mathbf{K}_j$, and $\xi_j$ denotes the softmax over all $j$ \cite{velivckovic2017graph}.
HGT uses \textit{sum} aggregation, followed by an \textit{update} step applying a nonlinearity $\sigma$ and a linear mapping to the aggregated vector.

\begin{figure}[htb]
\centering
\includegraphics[width=1\linewidth]{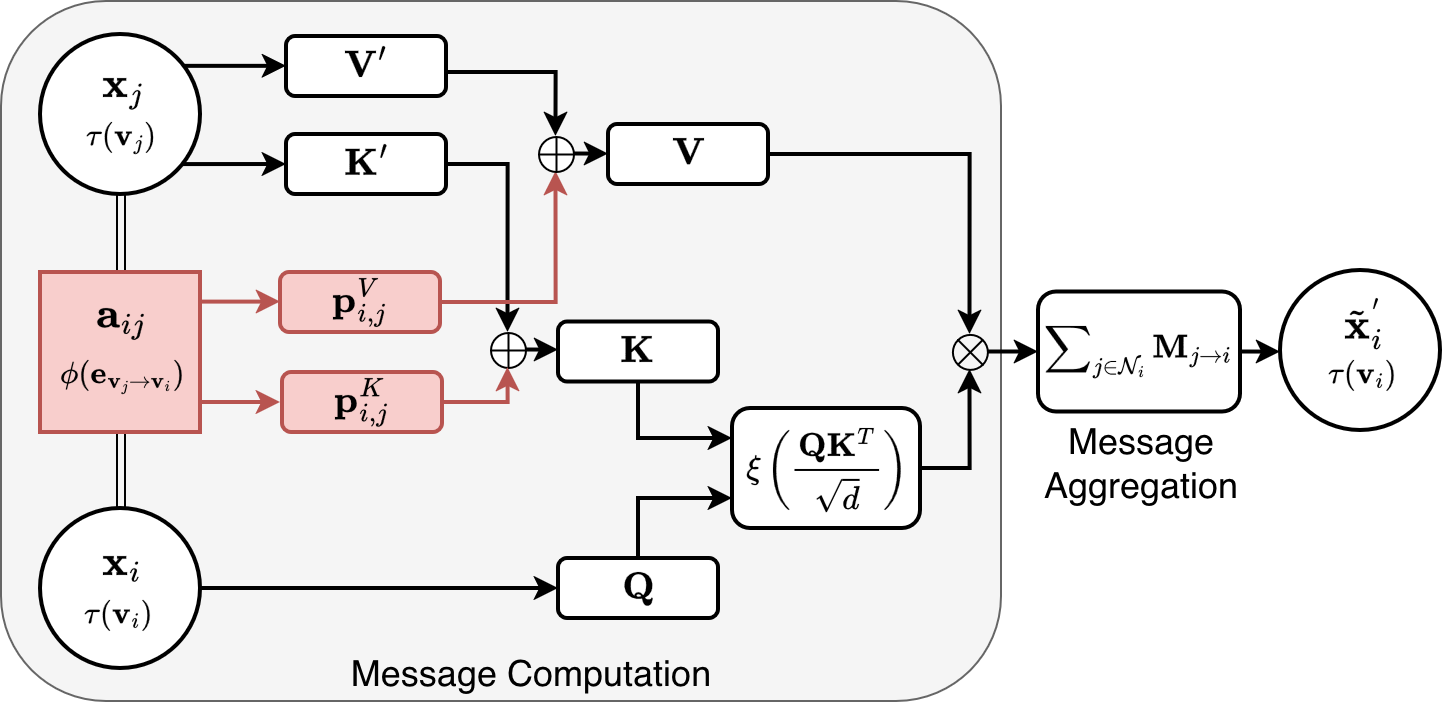}
\caption{
ART computes messages for each pair of connected \textit{source} and \textit{destination} nodes $\mathbf{v}_j$ and $\mathbf{v}_i$ in the input  graph, incorporating their relative affinity $\mathbf{a}_{i, j}$ into each message. Messages are aggregated into updated destination node vectors, $\tilde{\mathbf{x}}'_i$. Our novel relative affinity embeddings are highlighted in \textcolor{myred}{red}.}
\label{fig:art}
\end{figure}

ART is our graph transformer that injects pairwise relational features directly into self-attention, replacing relative position encodings \cite{shaw2018self, wu2021rethinking}, which use free learned 1D/2D embeddings, with $d$-dimensional embeddings of \textit{arbitrary} relationship vectors. As shown in \cref{fig:art}, we extend HGT using two independent encoders 
to embed the relative affinity $\mathbf{a}_{i,j}$ into \textit{key} and \textit{value} embeddings, 
implemented as a linear projection followed by \textit{BatchNorm}, 
\textit{ReLU} and another linear projection:
\begin{equation}
\label{eq:MLP_K}
\mathbf{p}_{i,j}^{K} = 
\max(0, \text{BN}(\mathbf{a}_{i,j}\mathbf{W}_1^K + \mathbf{b}_1^K))\mathbf{W}_2^K,
\end{equation}
\begin{equation}
\label{eq:MLP_V}
\mathbf{p}_{i,j}^{V} = 
\max(0, \text{BN}(\mathbf{a}_{i,j}\mathbf{W}_1^V + \mathbf{b}_1^V))\mathbf{W}_2^V.
\end{equation}
We use the \textit{key} embedding $\mathbf{p}_{i,j}^{K}$ to update the \textit{key} vector:
\begin{equation}
\mathbf{K}_j = \left(\mathbf{x}_j\mathbf{W}_K^{\tau} + \mathbf{b}_K^{\tau}\right) \mathbf{W}_K^{\phi} {+ \mathbf{p}_{i,j}^{K}},
\label{eq:updated_key}
\end{equation}
and the \textit{value} embedding $\mathbf{p}_{i,j}^{V}$ to update the \textit{value} vector:
\begin{equation}
\mathbf{V}_j=\left(\mathbf{x}_j\mathbf{W}_V^{\tau} + \mathbf{b}_V^{\tau}\right)\mathbf{W}_V^{\phi} {+ \mathbf{p}^V_{i,j}}.
\label{eq:updated_val}
\end{equation}
The \textit{aggregation} operator used in ART is the \textit{sum} operator:
\begin{equation}
\tilde{\mathbf{x}}'_i = \sum_{j \in \mathcal{N}_i} \xi_j \left( \frac{\mathbf{Q}_i\mathbf{K}_j^T}{\sqrt{d}} \right) \mathbf{V}_j,
\label{eq:agg}
\end{equation}
using the updated \textit{key} and \textit{value} vectors as defined in \cref{eq:updated_key,eq:updated_val}.
The \textit{update} step converts the aggregated vectors $\tilde{\mathbf{x}}'_i$ into $\mathbf{x}'_i$, defined as part of the ART block below.

\paragraph{ART Block.}
Following the pre-normalisation (Pre-LN) Transformer design \cite{xiong2020layer}, each ART block consists of LayerNorm, ART attention, a second LayerNorm, and a two-layer feed-forward network (FFN) (see Supplementary \cref{fig:artblock}). We use node-type-specific gated residuals, defined as $\mathbf{y}=\lambda^\tau \mathbf{u} + (1-\lambda^\tau)\mathbf{h}$ for both the ART and FFN skip connections, with $0\leq\lambda^\tau_{\text{ART}},\lambda^\tau_{\text{FFN}}\leq1$. Here, $\mathbf{h}$ is the sub-layer input, $\mathbf{u}$ is the intermediate ART or FFN output (e.g., the aggregated vector $\tilde{\mathbf{x}}'_i$), and $\mathbf{y}$ is the updated representation (e.g., $\mathbf{x}'_i$). The Graph Processor is formed by stacking $L$ ART blocks, and its output is passed to the ODN (\cref{ref:mdn}).

\subsection{Object Density Network}
\label{ref:mdn}

The updated node feature vectors from the $L$-th ART block for the last-timestep nodes $\mathcal{V}_T$ are processed by the \textit{ODN}, our adaptive mixture-density head outputting a 2D \textit{GMM} for the next gaze position at timestep $T+1$. The GMM has $K=|\mathcal{V}_{T}|$ components, one per node $\mathbf{v}_k\in\mathcal{V}_T$, so mixture capacity grows with scene complexity. For each node, a heterogeneous linear layer maps its updated feature vector $\mathbf{x}'_k$ to the component parameters $[\Delta \hat{x}_k, \Delta \hat{y}_k, \hat{\sigma}_{x_k}, \hat{\sigma}_{y_k}, \hat{\rho}_k, \hat{\pi}_k]$. We obtain valid parameters by softmaxing $\hat{\pi}_k$ so $0\leq\pi_k\leq1$ and $\sum \pi_k = 1$, bounding correlations with $\rho_k = \tanh(\hat{\rho}_k)$, and enforcing positive standard deviations via $\boldsymbol{\sigma}_k = \exp{([\hat{\sigma}_{x_k}, \hat{\sigma}_{y_k}])}$. The mean $\boldsymbol{\mu}_k$ is the node image-plane position $\boldsymbol{\mu}^0_k$ plus an offset $\Delta\boldsymbol{\mu}_k=[\Delta x_k, \Delta y_k]$, constrained by $\Delta\boldsymbol{\mu}_k=\Delta_{\text{max}}\tanh(\Delta\hat{\boldsymbol{\mu}}_k)$ with $\Delta_{\text{max}}=0.05$ (except structure node). High $\pi_k$ on the gaze node favours fixation maintenance; high $\pi_k$ on environment nodes favour attentional shifts to the corresponding objects or drivable regions. The future gaze distribution is then:

\begin{equation}
p\left((x, y)\right)= \sum_{k=1}^K\pi_k\mathcal{N}((x, y)|\boldsymbol{\mu}_k, \boldsymbol{\sigma}_k, \rho_k).
\end{equation}

\subsection{Training Objective}
\label{ref:mdn_loss}
Given a batch of $n$ spatio-temporal heterogeneous scene graphs, each representing a traffic scene over $T$ timesteps, and the corresponding ground truth future gaze positions, we train our model using the negative log likelihood loss:
\begin{equation}
\label{eq:loss}
\mathcal{L}_{\text{NLL}}= -\frac{1}{n}\sum_{i}^n \log \sum_{k=1}^K\pi_k\mathcal{N}(\mathbf{g}^{GT}_i|\boldsymbol{\mu}_k, \boldsymbol{\sigma}_k, \rho_k),
\end{equation}
where $\mathbf{g}^{GT}_i$ is the ground truth future gaze position for the $i$-th sample, and $\pi_k$, $\boldsymbol{\mu}_k$, $\boldsymbol{\sigma}_k$ and $\rho_k$ are the predicted parameters of the $k$-th Gaussian component.

\subsection{Simulating Gaze} 
The trained model can generate raw simulated gaze sequences by repeatedly estimating the future gaze position distribution from an input spatio-temporal scene graph $\mathbf{G}_T$, constructed from $T$ input frames. A random point is sampled from this distribution as the gaze position $\hat{g}_{T+1}$, which is then used to update the scene graph $\mathbf{G}_{T+1}$ for the next timestep. This process continues iteratively to produce gaze estimates for $\hat{g}_{T+2}$ and beyond. 

\section{Focus100 Dataset}

\label{ref:dataset}
\paragraph{Design.}For training and validating our methods we collected a large scale in-lab gaze dataset called \textit{Focus100}, from \(N=30\) participants (14M/16F; mean age \(36.9\) years, SD \(= 6.7\)), while they viewed hazardous egocentric driving videos. The dataset includes 100 egocentric driving videos, each \qty{60}{\second} long recorded at \qty{10}{\fps}, accompanied by synchronised \qty{60}{\hertz} gaze data. 
All participants reported driving frequently, minimally within the last week. Each participant viewed 30 sequences, with each video shown to 7--12 participants. The participants were seated \qty{57}{\cm} from a \qty{24}{\inch} desktop monitor, fitted with a \qty{60}{\hertz} \textit{Tobii Pro Nano} eyetracker, and viewed driving videos on the monitor while their gaze was recorded. During the viewing, in order to engage the participants in a  proxy task for driving \cite{xia2019predicting}, they undertook a `\textit{Hazard Perception Test}', modelled after the UK theory driving test \cite{govuk_hazard_perception}. We divide the 100 sequences into \textit{training} (70), \textit{validation} (10), and \textit{test} (20) sets. The details of Focus100, including ethical considerations, collection procedures, cross-dataset statistics, and the dataset release format, are given in the supplementary materials. 

\begin{figure}[t]
  \centering
  \includegraphics[width=0.94\linewidth]{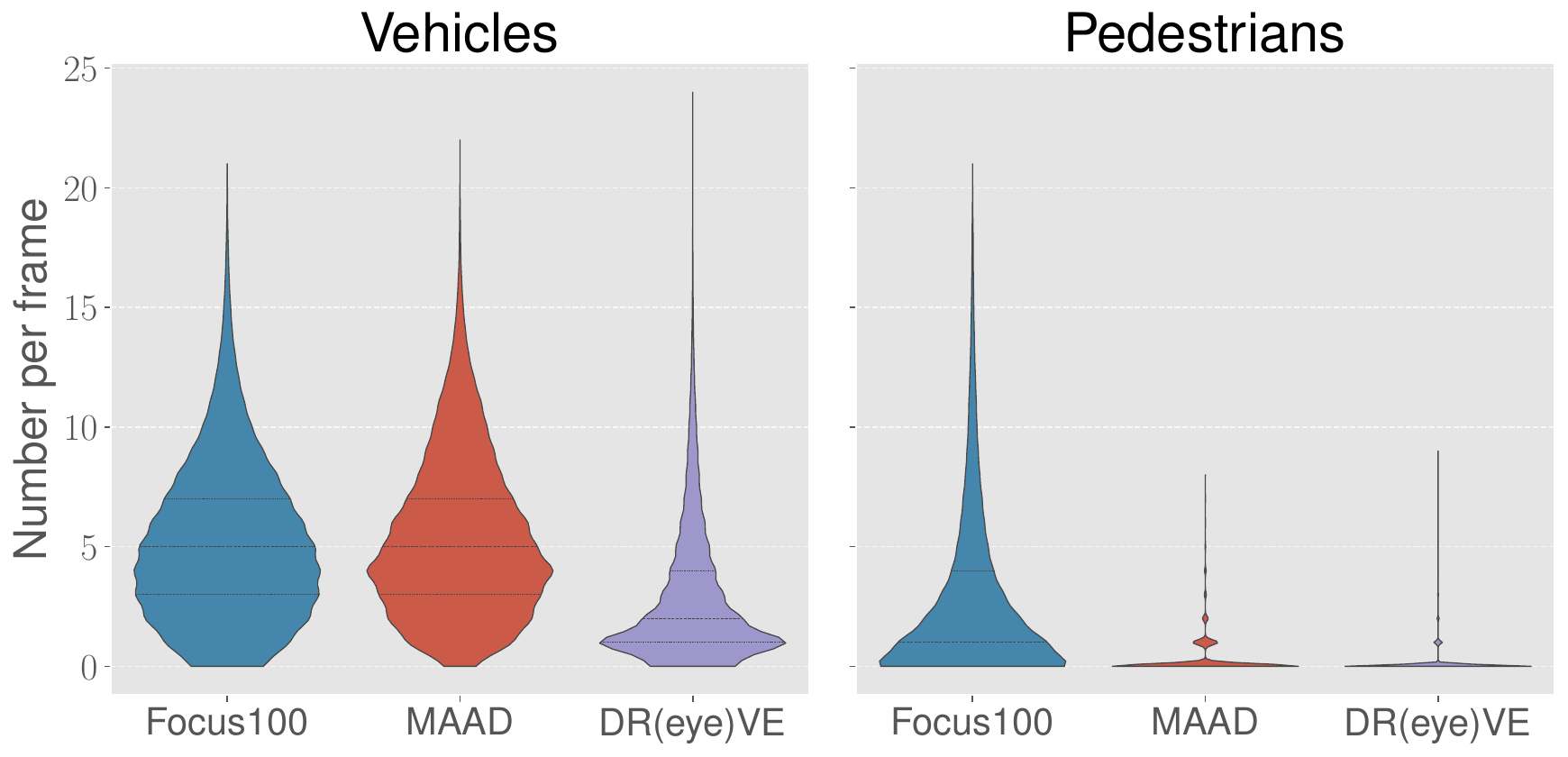}
   \caption{Violin plots of vehicle (left) and pedestrian (right) count per-frame in the Focus100, MAAD, and DR(eye)VE datasets.}
   \label{fig:MAADvF100}
\end{figure}

\paragraph{Driving Footage.} Driving footage was egocentric video from a front-facing, windscreen-mounted camera from a self-driving perception stack (\ang{52} field of view, \qty{10}{\fps}, at \numproduct{1280 x 806}~\si{\px}) resolution, calibrated to remove distortion and cropped to \numproduct{1280 x 640}~\si{\px} to exclude artefacts and the bonnet. Data were collected over two weeks around Brussels and Leuven (up to 8 h/day), covering urban, suburban, and highway scenes in daylight. From this corpus, we selected 100 one-minute clips to maximise variability: randomly sampled segments were scored by traffic density (vehicle/pedestrian detections), and 20 clips were chosen from each density quintile. See \cref{fig:MAADvF100} for a diversity comparison to the MAAD \cite{gopinath2021maad} and DR(eye)VE \cite{palazzi2018predicting} datasets; Focus100 displays greater variability in traffic conditions, especially with regard to pedestrian density, while including over twice the driving footage and over three times the gaze data of MAAD by duration. 

\paragraph{Hazard Annotation.}To facilitate area-of-interest analyses, annotators labelled and tracked bounding boxes of objects across all videos that met the definition of a hazard according to the UK driving theory test, where a hazard is defined as `something that would cause you to take action' \cite{govuk_hazard_perception}. The annotations were created using the CVAT annotation tool \cite{CVAT_ai_Corporation_Computer_Vision_Annotation_2023}, and each hazard was assigned a type and severity level. On average, $\mu = 4.08$ ($\sigma = 2.39$) hazards were annotated per \qty{60}{\second} sequence and tracked for an average of \qty{5.02}{\second}. In total, 207 hazards were labelled as low severity (`be ready to act') and 201 as severe (`take immediate evasive action'). Among these, 201 hazards involved pedestrians, 203 involved vehicles, and 4 involved `other' objects (\eg a dog). Hazard labels were admitted by consensus among three annotators.
\section{Experiments}
Here we present the experimental results comparing the performance of the proposed \textit{ART} module to other state-of-the-art gaze and attention estimation approaches. Experiments were carried out on our Focus100 dataset and the smaller and less diverse MAAD dataset \cite{gopinath2021maad}, the only other dataset which supplies synchronised raw gaze with driving footage.

\subsection{Experimental Setup}
\label{sec:experimental_setup}
\paragraph{Scene Graph Construction.}
Objects of the classes listed in \cref{tab:node_types} are detected with \textit{YOLOv8x} \cite{Jocher_Ultralytics_YOLO_2023} and mapped to their corresponding node types. Per detection, we extract appearance features from the 12th layer of a pretrained \textit{vgg16\_bn} network \cite{simonyan2014very} using \textit{ROIAlign} \cite{he2017mask}, yielding a 128-D appearance vector. The \textit{structure} node is obtained by estimating the drivable-area mask with \textit{YOLOPv2} \cite{han2022yolopv2}, resizing the mask to \numproduct{16 x 8}~\si{\px}, and flattening. We estimate depth with \textit{monodepth2} \cite{godard2019digging} and assign to each object node the mean of inverse disparity within its bounding box.

For Focus100, scene graphs span \qty{20}{} timesteps (\qty{1}{\second}), with directed temporal edges using offsets $\mathcal{T}_d={1,2,4,8,16}$ such that each node at time $t$ connects to nodes at $t-\Delta t$ for $\Delta t \in \mathcal{T}_d$, capturing multi-scale temporal context efficiently. Raw video (\qty{10}{\fps}) is upsampled to \qty{20}{\fps} by duplicating frames, while gaze is downsampled to \qty{20}{\hertz} so each timestep has synchronised object, structure, and gaze nodes. For MAAD (native \qty{25}{\fps} video), we downsample gaze to \qty{25}{\hertz} and build \qty{25}{}-timestep (\qty{1}{\second}) graphs with the same temporal connectivity. For both datasets, gaze is preprocessed by linear interpolation across blinks at the native sampling rate (see Supplementary for full details).

\paragraph{Training.}
\label{ref:training}
We use the Adam optimiser \cite{kingma2014adam} and a batch size of $128$ to optimise the loss defined in \cref{eq:loss} for $50$ epochs on 4 \textit{NVIDIA L40S} GPUs with \textit{float16} precision. We train with a base learning rate of \num{3e-4} on Focus100 and \num{1e-3} on MAAD, with weight decay \num{1e-6}. The ODN head uses \num{0.1}$\times$ the base learning rate. The best model is chosen as the checkpoint achieving the minimum validation loss.

\begin{table*}[htbp]
\centering
\caption{Sequence, dynamics and saliency map metrics for different approaches on Focus100 and MAAD. Arrows indicate the direction of improvement; best score per metric is marked in \textbf{bold}, second is \underline{underlined}. DTW and LEV are in thousands and are sensitive to sequence length, hence the difference in MAAD and Focus100 scores. Dynamics metrics closest to human statistics follow the same notation. Gaze sequences generated by Itti, GBVS and Gaussian  produced insufficient fixations to include in the dynamics comparison. Standard deviation across sequences is reported for sequence metrics and across fixations for dynamics metrics.}
\label{tab:metrics}
\resizebox{\textwidth}{!}{%
\begin{tabular}{@{}clcccccccccc@{}}
\toprule
& & \multicolumn{3}{c}{\textbf{Sequence}} & \multicolumn{3}{c}{\textbf{Dynamics}} & \multicolumn{3}{c}{\textbf{Saliency}} \\
\cmidrule(lr){3-5} \cmidrule(lr){6-8} \cmidrule(lr){9-11}
\textbf{Dataset} & \textbf{Model} & \textbf{TC $\uparrow$} & \textbf{DTW $\downarrow$} & \textbf{LEV $\downarrow$} & \textbf{Fix Dur (s)} & \textbf{Fix Rate (fix/s)} & \textbf{AOI TFF (s)} & \textbf{NSS $\uparrow$} & \textbf{IG $\uparrow$} & \textbf{AUC $\uparrow$} \\ 
\midrule

\multirow{9}{*}{\rotatebox[origin=c]{90}{\textbf{Focus100}}}

& \textit{Human} & \textit{0.46 $\pm$ 0.09} & \textit{30.93 $\pm$ 3.71} & \textit{1.03 $\pm$ 0.09} & \textit{0.44 $\pm$ 0.56} & \textit{1.61 $\pm$ 0.00} & \textit{0.41 $\pm$ 0.56} & - & - & - \\

& Gaussian & 0.03 $\pm$ 0.01 & 66.22 $\pm$ 4.78 & 1.55 $\pm$ 0.04 & -  & - & - & 3.375 & 8.831 & 0.918 \\

& Itti \cite{itti1998model} & 0.03 $\pm$ 0.02 & 157.08 $\pm$ 3.13 & 2.03 $\pm$ 0.01 & - & - & - & 0.321 & 6.455 & 0.606 \\

& GBVS \cite{harel2006graph} & 0.03 $\pm$ 0.01 & 133.18 $\pm$ 2.37 & 1.98 $\pm$ 0.01 & -  & - & - & 1.333 & 7.220 & 0.849 \\

& GLC \cite{lai2024eye} & 0.16 $\pm$ 0.05 & \underline{48.72 $\pm$ 12.77} & \underline{1.24 $\pm$ 0.12} & \underline{0.21 $\pm$ 0.15} & \underline{0.57 $\pm$ 0.00} & \underline{1.72 $\pm$ 1.48} & \underline{4.481} & 9.162 & 0.914 \\

& DReyeVENet \cite{palazzi2018predicting} & \underline{0.23 $\pm$ 0.09} & 49.23 $\pm$ 5.04 & 1.43 $\pm$ 0.04 & 0.12 $\pm$ 0.04 & 0.07 $\pm$ 0.00 &  2.77 $\pm$ 1.84 & 3.749 & 9.041 & 0.920 \\

& SCOUT \cite{kotseruba2024understanding} & 0.22 $\pm$ 0.08 & 51.76 $\pm$ 7.40 & 1.45 $\pm$ 0.06 & 0.12 $\pm$ 0.03 & 0.05 $\pm$ 0.00 & 3.62 $\pm$ 2.29 & 4.152 & 9.440 & 0.933 \\

& ViNet \cite{jain2021vinet} & \textbf{0.23 $\pm$ 0.06} & 49.68 $\pm$ 5.83 & 1.41 $\pm$ 0.05 & 0.12 $\pm$ 0.04 & 0.07 $\pm$ 0.00 &  2.90 $\pm$ 1.97 & 4.310 & \underline{9.471} & \underline{0.938} \\

& \textbf{ART} (ours) & 0.22 $\pm$ 0.05 & \textbf{42.31 $\pm$ 4.88} & \textbf{1.23 $\pm$ 0.10} & \textbf{0.41 $\pm$ 0.40} & \textbf{1.64 $\pm$ 0.00}  & \textbf{0.59 $\pm$ 0.47} & \textbf{4.864} & \textbf{9.728} & \textbf{0.945} \\

\midrule

\multirow{9}{*}{\rotatebox[origin=c]{90}{\textbf{MAAD} \cite{gopinath2021maad}}}

& \textit{Human} & \textit{0.42 $\pm$ 0.15} & \textit{2.65 $\pm$ 1.13} & \textit{0.10 $\pm$ 0.03} & \textit{0.53 $\pm$ 0.57} & \textit{1.25 $\pm$ 0.00} & - & - & - & - \\

& Gaussian & 0.14 $\pm$ 0.03 & 6.15 $\pm$ 0.95 & 0.20 $\pm$ 0.01 & - & - & - & 3.452 & 9.327 & 0.939 \\

& Itti \cite{itti1998model} & 0.15 $\pm$ 0.03 & 16.21 $\pm$ 0.98 & 0.25 $\pm$ 0.00 & - & - & - & 0.240 & 6.607 & 0.607 \\

& GBVS \cite{harel2006graph} & 0.14 $\pm$ 0.02 & 13.63 $\pm$ 0.70 & 0.25 $\pm$ 0.00 & - & - & - & 1.467 & 7.503 & 0.876 \\

& GLC \cite{lai2024eye} & \underline{0.23 $\pm$ 0.10} & \underline{3.25 $\pm$ 1.45} & \underline{0.14 $\pm$ 0.02} & \underline{0.18 $\pm$ 0.09} & \underline{0.56 $\pm$ 0.00} & - & 4.904 & 8.900 & 0.869 \\

& DReyeVENet \cite{palazzi2018predicting} & 0.20 $\pm$ 0.08 & 4.69 $\pm$ 1.68 & 0.16 $\pm$ 0.02 & 0.14 $\pm$ 0.04 & 0.09 $\pm$ 0.00 & - & 4.754 & 9.233 & 0.936 \\

& SCOUT \cite{kotseruba2024understanding} & 0.19 $\pm$ 0.08 & 5.91 $\pm$ 1.32 & 0.18 $\pm$ 0.01 & 0.13 $\pm$ 0.02 & 0.01 $\pm$ 0.00 & - & 4.191 & 9.735 & \underline{0.952} \\

& ViNet \cite{jain2021vinet} & 0.20 $\pm$ 0.09 & 4.20 $\pm$ 1.60 & 0.16 $\pm$ 0.02 & 0.15 $\pm$ 0.05 & 0.15 $\pm$ 0.00 & - & \textbf{5.733} & \textbf{10.264} & 0.949 \\

& \textbf{ART} (ours) & \textbf{0.46 $\pm$ 0.11} & \textbf{2.70 $\pm$ 1.22} & \textbf{0.10 $\pm$ 0.04} & \textbf{0.70 $\pm$ 0.68} & \textbf{0.99 $\pm$ 0.00} & - & \underline{4.926} & \underline{9.778} & \textbf{0.953} \\

\bottomrule
\end{tabular}
}
\end{table*}

\paragraph{Simulation.}
We use our trained models to simulate gaze sequences matched to each human ground-truth sequence. Starting with the initial 20 (ART) or 25 (MAAD) timesteps of each ground-truth sequence, we iteratively sample from the predictive ODN distribution to generate subsequent spatio-temporal graphs and predictive distributions. These simulated sequences are labelled \textit{ART} in Table \ref{tab:metrics}.
Saliency maps were generated by running 50 simulations per sequence with random initialisation, detecting fixations with the EyeMMV algorithm \cite{krassanakis2014eyemmv}, and convolving fixation maps per frame with a Gaussian kernel following \cite{xu2014predicting}.

\paragraph{Metrics.}

We evaluate our method across three domains: raw gaze sequences, saliency maps, and scanpath dynamics. For raw gaze sequences, we compare generated sequences to human data using three time-series metrics focused on temporal alignment: Dynamic Time Warping (\textit{DTW}; \cite{berndt1994using}), Temporal Correlation (\textit{TC}; \cite{shepherd2010human}), and Levenshtein distance (\textit{LEV}; \cite{fahimi2021metrics}). Each generated sequence is paired with its closest ground-truth match, and the average of these best matches provides an overall score. For a human baseline, we apply this procedure in a leave-one-out setting, pairing each human sequence with its closest match among the remaining human sequences and averaging these best matches.

For scanpath dynamics, we apply the EyeMMV fixation filter \cite{krassanakis2014eyemmv} to extract fixation positions and durations. We compute mean fixation duration (\textit{Fix Dur}), fixation rate (\textit{Fix Rate}), 
and time-to-first-fixation within a defined radius of an AOI's centre-of-mass (\textit{AOI TFF}; we set the threshold radius as 10\% of image width). To evaluate saliency maps, we use three common metrics \cite{kummerer2018saliency}: Normalised Scanpath Saliency (\textit{NSS}), Information Gain (\textit{IG}), and Area Under the Curve (\textit{AUC}).

\paragraph{Baselines.}
Our main comparisons are with approaches which estimate a spatial probability of gaze (\ie a saliency map) given video input, namely \textit{Global-Local Correlation (GLC)} \cite{lai2024eye}, \textit{SCOUT} \cite{kotseruba2024understanding}, \textit{ViNet} \cite{jain2021vinet}, and \textit{DReyeVENet} \cite{palazzi2018predicting}; 
we used publicly available official implementations of these models.
GLC is the current state-of-the-art in egocentric gaze estimation from video, SCOUT achieves state-of-the-art performance for driver gaze prediction on DR(eye)VE and BDD-A \cite{xia2019predicting} datasets (we use their task-free variant to align with our method and other baselines), ViNet is a recent model for video saliency prediction, while DReyeVENet serves as a well-established baseline for driver gaze prediction.
We also include Itti's method (\textit{Itti}) \cite{itti1998model}, graph-based visual saliency (\textit{GBVS}) \cite{harel2006graph}, and a 2D Gaussian distribution fitted across all fixations in the training set (\textit{Gaussian}). To train the baselines relying on aggregated saliency maps, we computed a ground-truth saliency map for each frame in the datasets by convolving a Gaussian over fixation locations across subjects. For evaluating saliency estimation methods on sequence and dynamics metrics, we produce gaze sequences by sampling a gaze position per frame proportionally to each frame's predicted gaze distribution.

\subsection{Quantitative Results}

\cref{tab:metrics} demonstrates that ART matches or outperforms all baselines in raw gaze and scanpath generation across both Focus100 and MAAD datasets, and approaches human-level results on several metrics. On Focus100 we also note improvements in saliency map generation across all metrics, which is notable given that several baselines are specifically tailored for this task. These results emphasise ART's versatility in handling dynamic visual scenes and accurately estimating gaze behaviour across different temporal scales.

\begin{figure*}[ht]
\centering
\includegraphics[width=\linewidth]{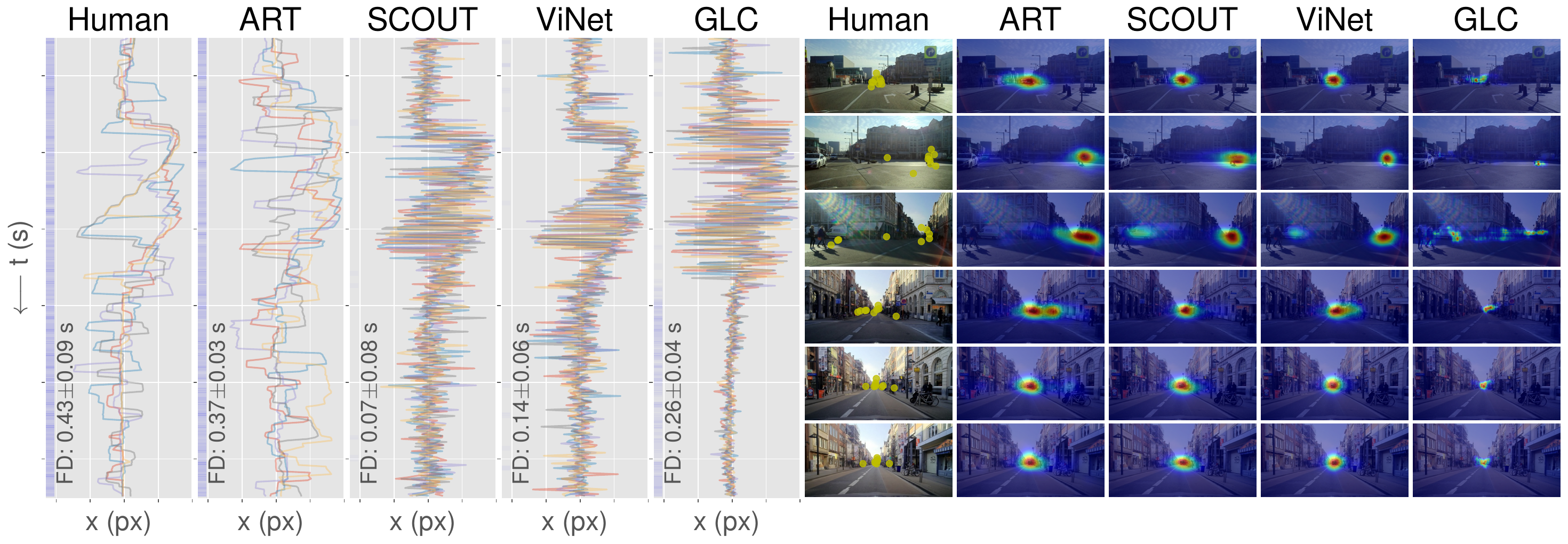}
\caption{ Gaze sequences and saliency maps generated on a \qty{15}{\s} clip of Focus100. The first column shows human gaze sequences, followed by those generated by ART, SCOUT, ViNet and GLC models. Each trace represents a single simulation, with the y-axis indicating time and x-axis showing left-to-right gaze position; blue marks to the left show detected fixations, and average fixation duration (FD) per method is given. On the right, we display observed fixations for humans and model-generated saliency maps for the same video frames, temporally aligned with the gaze sequences for direct comparison. See the Supplementary for further examples.}
\label{fig:main_visualisation}
\end{figure*}

\subsection{Qualitative Results}

Our qualitative analysis (\cref{fig:main_visualisation}) reinforces the quantitative results previously reported: gaze sequences generated by ART exhibit a striking resemblance to human gaze behaviour when compared to the outputs from baselines. ART sequences closely mirror human temporal dynamics, with fixation frequency and fixation duration that match ground-truth data. The blue fixation markers in \cref{fig:main_visualisation} further highlight this difference: while ART produces contiguous fixation periods, the baselines generate few, if any. Moreover, the variance of gaze paths produced by ART matches the variability seen in human observers, in contrast to the mean-biased trajectories with residual noise produced by the other methods (see Supplementary for analysis of this variance). Without explicit saliency supervision, unlike the baselines, ART produces saliency maps consistent with human fixations, indicating that modelling raw gaze dynamics captures the underlying attention structure.

\subsection{Ablation Study}

In an ablation study, we vary the number of historical timesteps in the spatio-temporal graph, replace ART with HGT \cite{hu2020heterogeneous} or HEAT \cite{mo2021heterogeneous}, and substitute ODN with a standard MDN head. We simulate gaze sequences as in the main experiment and report sequence-level metrics. \textbf{Temporal window size}: Smaller $T$ generally reduces alignment with human sequences (\cref{tab:ablation}), indicating that longer temporal context helps capture dependencies. \textbf{Graph processor}: Using HGT or HEAT in place of ART lowers sequence plausibility, highlighting the importance of ART’s relational edge modelling. \textbf{Object-based distribution}: Replacing ODN with an MDN significantly degrades performance; unlike ODN, the standard MDN is not object-conditioned and cannot adapt mixture capacity to scene complexity.

\begin{table}[htbp]
\centering
\caption{Ablation study across temporal window, graph processor, and output distribution head variants. ART with 20 timesteps and ODN head is our reported method elsewhere.}
\label{tab:ablation}
\resizebox{\columnwidth}{!}{%
\begin{tabular}{@{}lllccc@{}}
\toprule
\textbf{Processor} & \textbf{Time} & \textbf{Head} & \textbf{TC $\uparrow$} & \textbf{DTW $\downarrow$} & \textbf{LEV $\downarrow$} \\
\midrule

\textit{ART} & \textit{20} & \textit{ODN} & \textit{0.22 $\pm$ 0.05} & \textit{42.31 $\pm$ 4.88} & \textit{1.23 $\pm$ 0.10} \\
\midrule

HGT \cite{hu2020heterogeneous} & 20 & ODN & 0.21 $\pm$ 0.06 & 42.72 $\pm$ 5.17 & 1.28 $\pm$ 0.09 \\
HEAT \cite{mo2021heterogeneous} & 20 & ODN & 0.13 $\pm$ 0.05 & 59.50 $\pm$ 7.17 & 1.47 $\pm$ 0.08 \\
\midrule

ART & 20 & MDN$_{k=10}$ & 0.14 $\pm$ 0.04 & 44.78 $\pm$ 4.80 & 1.30 $\pm$ 0.10 \\
ART & 20 & MDN$_{k=20}$ & 0.14 $\pm$ 0.04 & 45.69 $\pm$ 3.51 & 1.32 $\pm$ 0.09 \\
\midrule

ART & 8 & ODN  & 0.17 $\pm$ 0.04 & 43.46 $\pm$ 4.50 & 1.26 $\pm$ 0.09 \\
ART & 1 & ODN  & 0.17 $\pm$ 0.07 & 42.35 $\pm$ 5.24 & 1.24 $\pm$ 0.09 \\
\bottomrule
\end{tabular}
}
\end{table}

\section{Conclusion and Limitations}

In this work, we introduced a novel dynamical systems approach to unify attention modelling in dynamics scenes that significantly improves alignment with human gaze behaviour; demonstrating strong performance across gaze sequence generation, scanpath dynamics, and saliency map quality. Our framework and the release of the Focus100 dataset open avenues for further research in temporal gaze modelling. We note three limitations: Focus100 was collected in a controlled laboratory setting rather than on-road (see Supplementary for detailed treatment of this point); ART relies on an upstream perception stack, where failures can propagate to gaze prediction; and we do not explicitly model driver intent, which can modulate attention \cite{kotseruba2024scout+}.

\clearpage
\setcounter{page}{1}
\maketitlesupplementary

\section{The Focus100 Dataset}

Focus100 is a new dataset designed to facilitate research on dynamic human attention in driving scenarios, particularly for the development and evaluation of gaze estimation models. This dataset addresses critical limitations in existing driving gaze datasets, which often lack raw gaze data or sufficient scenario diversity \cite{fang2021dada, xia2019predicting, deng2019drivers, deng2016does}.  Unlike datasets that provide only aggregated saliency maps, Focus100 provides high-resolution, time-stamped gaze sequences from 30 participants viewing 100 egocentric driving videos. This rich data enables the study of fine-grained temporal attention patterns and scanpath dynamics, crucial for understanding human behaviour in complex driving environments.

Although the DR(eye)VE dataset \cite{palazzi2018predicting} offers raw gaze sequences, it is hampered by limitations such as temporal misalignment, low scenario complexity, having only a single gaze sequence recorded per driving video, and lack of gaze data in the image plane (instead registered to moving driver-worn eye tracker glasses) \cite{kotseruba2024data}. While efforts have been made to enrich the dataset through in-lab gaze tracking \cite{gopinath2021maad}, these have only addressed a small subset of the data. Focus100 overcomes these shortcomings by providing a diverse set of driving scenarios, several precise gaze recordings per driving video (in image coordinates), making it a valuable resource for advancing research in driver attention and automotive safety.

\begin{figure*}[htbp]
    \centering
    \begin{tabular}{ccc}
        \includegraphics[width=0.3\textwidth]{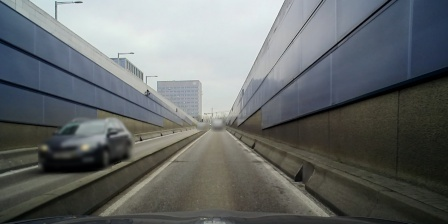} & 
        \includegraphics[width=0.3\textwidth]{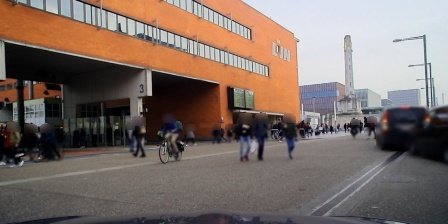} & 
        \includegraphics[width=0.3\textwidth]{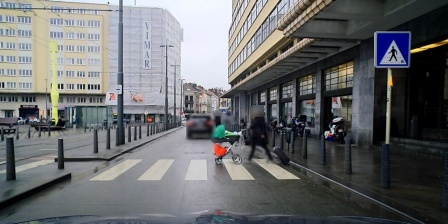} \\[1em]
        \includegraphics[width=0.3\textwidth]{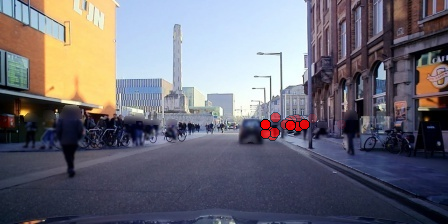} & 
        \includegraphics[width=0.3\textwidth]{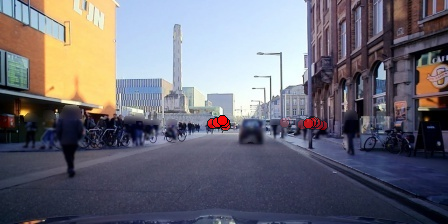} & 
        \includegraphics[width=0.3\textwidth]{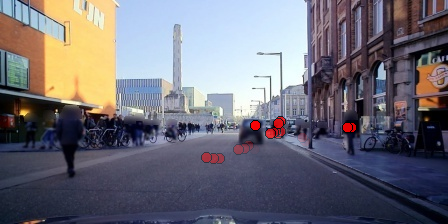}
    \end{tabular}
    \caption{Examples from Focus100. The top row shows diversity in pedestrian traffic, hazardousness, and road type. The bottom row shows the same video frame overlaid with the gaze samples of three separate subjects over the previous \qty{2}{\second} window, where higher alpha of the gaze position corresponds to a more recent sample. The example demonstrates the diversity of temporal gaze patterns across subjects for the same stimuli --- information which is lost through averaging in traditional saliency map data representations. }
    \label{fig:dataset_examples}
\end{figure*}

\subsection{Data Collection}

\paragraph{Driving Footage}
The driving videos incorporated in Focus100 were captured using a test vehicle equipped with a front-facing camera installed behind the windscreen to provide a point-of-view similar to that of the driver. The camera used for recording had a 52-degree horizontal visual angle and captured footage at 10 frames per second with a resolution of \numproduct{1280 x 806}~\si{\px}. Comparable FoV and frame-rate settings are used in several driving datasets, \eg Waymo Open \cite{sun2020scalability}: $55$$^{\circ}$@$10$Hz; BDD100K \cite{yu2020bdd100k}: $48$$^{\circ}$@$30$Hz; Euro-PVI \cite{bhattacharyya2021euro}: $52$$^{\circ}$@$10$Hz.
All images were subjected to a calibration process to eliminate distortion and cropped to \numproduct{1280 x 640}~\si{\px} resolution to remove calibration artefacts and ego-car bonnet pixels.

Driving sessions, which lasted up to 8 hours per day, were carried out over two weeks in and around the cities of Brussels and Leuven, Belgium. This geographical diversity allowed recording of a wide range of driving environments, including urban areas, suburban neighbourhoods, and highways. All video recordings were conducted during daylight hours to ensure good visibility.

From this collection of driving footage, a subset of 100 1-minute videos was selected to form the Focus100 dataset. The selection process aimed to maximise variance in driving complexity. To achieve this, we analysed randomly sampled 1-minute sections from the entire dataset and estimated the traffic density based on the total number of vehicle and pedestrian detections. We then selected 20 videos from each quintile of traffic density, ensuring a balanced representation of different traffic conditions within Focus100.

\paragraph{Gaze Data}
To study natural gaze behaviour in response to realistic driving scenarios, we designed an experiment that simulated the experience of driving while capturing participants' eye movements. This involved presenting participants with a series of engaging driving video clips and asking them to perform a hazard perception task, mirroring the hazard perception component of the UK driving test \cite{govuk_hazard_perception}. This task required participants to actively monitor the videos for potential hazards and respond by pressing the CTRL key whenever they perceived a developing hazard. This approach ensured that participants remained engaged and attentive while providing valuable insights into their natural gaze patterns in response to dynamic driving situations.

Thirty frequent drivers, 14 male and 16 female, with an age range between 21 and 60 years (M=36.9, SD=6.7), were recruited for this study. All participants had held a valid driver's license for at least three years, had normal vision, and confirmed that they had driven within the past week. Before commencing the study, each participant provided informed consent.

The study was conducted in a controlled laboratory setting. Participants were seated \qty{57}{\cm} from a \qty{24}{\inch} Dell P2423 monitor, with the freedom to slightly adjust their position for comfort. A Tobii Pro Nano eye tracker, attached to the lower edge of the monitor, recorded their gaze data at \qty{60}{\hertz}. Participants used a standard Logitech K120 keyboard to provide responses during the hazard perception task.

Before each session, the eye tracker was calibrated to ensure accurate gaze capture for each participant. The participants were then briefed on the purpose and procedure of the study, given practice on the hazard perception task to familiarise themselves with the response mechanism, and asked about their driving history.

During data collection, participants viewed a series of 1-minute egocentric driving video clips. Each participant viewed 30 unique clips and each clip was shown to 7--12 randomly assigned participants, ensuring a balanced representation of individual viewing patterns and responses across the dataset. The order of presentation of the clips was balanced to maintain participant engagement and minimise fatigue. Regular breaks were also incorporated into the session to further combat fatigue and ensure data quality. Due to technical issues during gaze recording, we omit 10 recordings from the dataset, leaving 890 1-minute gaze recordings across 30 subjects.

\paragraph{Hazard Annotations}
Three annotators labelled and tracked the bounding boxes of objects in the scene that met the definition of a hazard from the UK driving theory test \cite{govuk_hazard_perception}; \textit{A developing hazard is something that would cause you to take action, like changing speed or direction}. The objects were annotated using the \textit{CVAT} \cite{CVAT_ai_Corporation_Computer_Vision_Annotation_2023} annotation tool.
Each hazard was also assigned a \textit{type}: \textit{pedestrian}, \textit{vehicle}, \textit{other}; and a \textit{severity} level: \textit{low - preparing to act}, or \textit{high - take evasive action, \eg immediate application of the brakes}.
On average, 4.08 $\pm$ 2.39 hazards were annotated per \qty{60}{\second} sequence (sequences are diverse in hazard counts) and tracked an average for \qty{5.02}{\second}.
In total, 207 hazards were low severity and 201 severe; 201 hazards were pedestrians, 203 vehicles, and 4 `other' (\eg, a dog). The labels were accepted by consensus among three annotators.

\subsection{Ethics Statement}

From the onset, privacy and ethics standards were critical to this data collection effort. The study was conducted in strict accordance with GlimpseML and Toyota Motor Europe institutional research policies. Participants in the gaze collection were fully informed about the purpose, procedures, and potential risks of the study, including the intention to publish anonymised data for academic research purposes. They were given the opportunity to ask questions and were free to withdraw at any time without consequence. Participants also retained the right to redact their own data at any point before or after publication.

To protect the privacy of individuals in driving videos, all personally identifiable information (PII) has been carefully removed. All detected faces and license plates in the videos were automatically blurred to ensure that individuals and vehicles could not be identified; this was then manually checked frame-by-frame by three annotators. The gaze data provided in the dataset has been processed to remove any information that could potentially identify individual participants. All personal identifiers associated with the gaze data, such as participant names or ID numbers, gender, age, recording locations, and times have been removed.

The Focus100 dataset is stored securely in a GDPR-compliant manner on MFA-protected servers with restricted access within the EU to prevent unauthorised access and ensure data confidentiality.  The dataset is restricted to research or academic use only and requires institutional registration for access. Users of the dataset are expected to adhere to ethical research practices and comply with all relevant data privacy regulations, including GDPR. Commercial use is strictly prohibited.

By implementing these measures, we prioritise the privacy and anonymity of all individuals involved, while providing a valuable resource for the research community to advance the study of driver attention and automotive safety.

\subsection{Characteristics}

Focus100 comprises 100 egocentric driving videos, each 60 seconds in duration, captured at 10 frames per second with a resolution of \numproduct{1280 x 640}~\si{\px} and a 52$^\circ$ field of view. These videos encompass a diverse range of traffic conditions providing rich visual stimuli representative of real-world driving scenarios. See \Cref{tab:datasets} for the relevant statistics of the dataset.

Compared to the only comparable in-lab dataset, MAAD \cite{gopinath2021maad}, a small subset of the DR(eye)VE dataset \cite{palazzi2018predicting}, Focus100 offers significant advantages in terms of scale and diversity. With nearly 15 hours of gaze recordings from 30 participants, Focus100 surpasses MAAD's 4.83 hours of engaged gaze data, collected from 23 subjects across only 8 videos (all in urban downtown settings). This increased scale translates into a broader representation of driving situations.
The distribution of traffic complexity in our new dataset in comparison with the DR(eye)VE and MAAD datasets is shown in Fig 4 of the paper; Focus100 surpasses both in vehicle and pedestrian diversity.

\begin{figure}[t]
  \centering
  \includegraphics[width=0.94\linewidth]{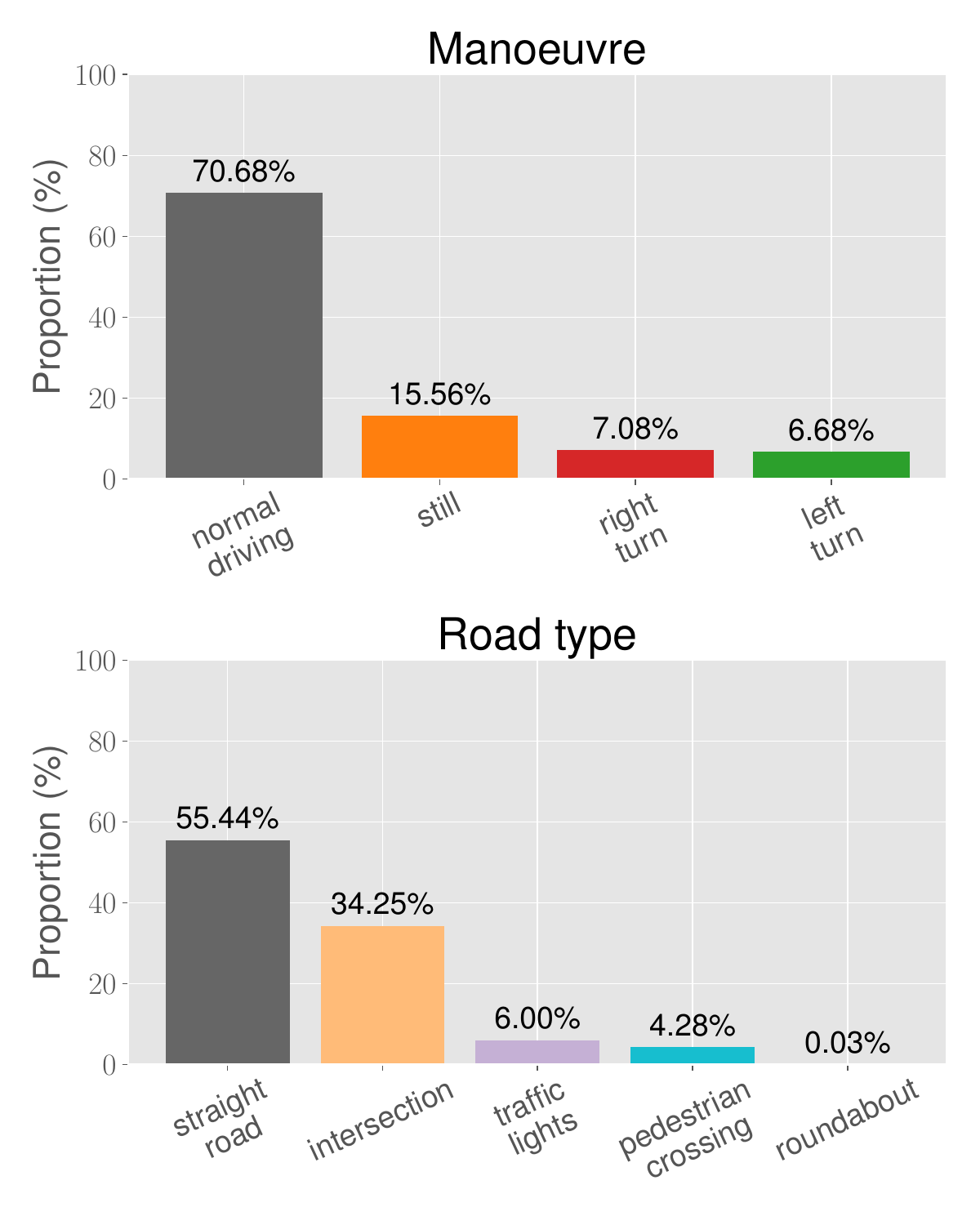}
   \caption{Frame-level distributions of ego-car manoeuvres and road types in the Focus100 dataset.}
   \label{fig:F100distributions}
\end{figure}

Following \cite{palazzi2018predicting}, we divide the manoeuvres of the ego-car into 4 classes: normal driving, turning left, turning right, and being still (defined as the vehicle being completely stationary or moving slowly). Each frame in the dataset was manually labelled with both the ego-car manoeuvre and the road type. The road types are divided into 5 categories: straight road, intersection, traffic lights, pedestrian crossing, and roundabout. These distributions are visualised in \Cref{fig:F100distributions}. 

\begin{table}[htbp]
\centering
\caption{Key statistics of MAAD and Focus100 gaze datasets. MAAD collected gaze over a subset of the DR(eye)VE dataset (8 videos in downtown/urban settings). Detections are presented per frame, with mean and standard deviation across the whole dataset. Note that these detections were estimated on downsampled \numproduct{448 x 224} and \numproduct{398 x 224} image resolutions on Focus100 and MAAD, respectively, matching the resolutions used in our methods.}
\label{tab:datasets}
\resizebox{\columnwidth}{!}{%
\begin{tabular}{@{}llcc@{}}
\toprule
\textbf{Category} & \textbf{Measure} & \textbf{MAAD \cite{gopinath2021maad}}* & \textbf{Focus100} \\
\midrule
Video & \# Videos & 8 & 100 \\
       & Video Length (\si{\second}) & 300 & 60 \\
       & Anonymised & no & yes \\
       & Resolution (\si{\px}) & \numproduct{1920 x 1080} & \numproduct{1280 x 640} \\
       & FoV ($^\circ$)  & unknown & 52 \\
       & Frequency (\si{\hertz}) & 25 & 10 \\
\midrule
Detections
       & Pedestrian & 0.38 $\pm$ 1.01 & 2.59 $\pm$ 3.26 \\
       & Traffic light & 0.34 $\pm$ 0.87 & 0.12 $\pm$ 0.48 \\
       & Stop sign & 0.02 $\pm$ 0.13 & 0.05 $\pm$ 0.22 \\
       & Car & 5.57 $\pm$ 3.36 & 3.17 $\pm$ 3.03 \\
       & Bicycle & 0.06 $\pm$ 0.32 & 1.26 $\pm$ 2.09 \\
       & Truck & 0.26 $\pm$ 0.52 & 0.37 $\pm$ 0.71 \\
       & Bus & 0.06 $\pm$ 0.24 & 0.18 $\pm$ 0.48 \\
       & Motorcycle & 0.02 $\pm$ 0.17 & 0.08 $\pm$ 0.35 \\
\midrule
Gaze   & \# Subjects & 23 & 30 \\
       & Recorded & in-lab & in-lab \\
       & Frequency (\si{\hertz}) & 250 & 60 \\
       & Sub/video (range) & 6--11 & 7--12 \\
       & Sub/video (avg) & 7.25 & 8.9 \\
       & Subject age & 20--55 & 21--60 \\
       & Subject gender & 22M--6F & 14M--16F \\
       & Total dur (hours) & 4.83 & 14.83 \\
       & Licensure (years) & $>$2 & $>$3 \\
\bottomrule
\end{tabular}
}
\\
\footnotesize{* MAAD collected data across several conditions with distractions or reduced visibility, here we report the statistics for the control condition.}
\end{table}

\subsection{Data Format Overview}

The Focus100 dataset comprises anonymised driving videos with associated viewer gaze. We also provide object detections extracted from the original videos (non-anonymous):

\begin{description}

\item[\textbf{Video:}]  100 \qty{60}{}-second videos at \qty{10}{\hertz} frame rate, anonymised to remove PII, cropped and downsampled to \numproduct{1280 x 640}~\si{\px}.

\item[\textbf{Gaze:}]  890 \qty{60}{}-second gaze sequences sampled at \qty{60}{\hertz}, mean of left and right eye gaze positions in image space, for at least 7 subjects per video. Each sample is synchronised and associated to a video frame.

\item[\textbf{Detections:}]  YOLOv8x \cite{Jocher_Ultralytics_YOLO_2023} detections per frame for the following classes: pedestrian, traffic light, stop sign, car, bicycle, truck, motorcycle. 

\end{description}

\subsection{Discussion}

Focus100 offers key advantages over existing driver attention datasets. It provides raw, temporally aligned gaze sequences for fine-grained visual attention analysis and covers diverse driving environments. Each of its 100 videos was viewed by 30 participants, yielding 7--12 gaze recordings per clip. All data collection followed strict privacy and ethical standards. We discuss potential limitations below.

\paragraph{Sampling Frequency} A potential concern is the video frame rate (\qty{10}{\fps}) and eye-tracker sampling rate (\qty{60}{\hertz}) used in Focus100. However, in practice, 10 frames per second is a standard in many autonomous driving datasets and perception stacks, for instance, the Waymo Open Dataset and Euro-PVI camera streams operate at \qty{10}{\hertz} \cite{sun2020scalability,bhattacharyya2021euro}, while nuScenes imagery is captured at \qty{12}{\hertz} \cite{caesar2020nuscenes}. This frame rate is sufficient to capture the temporal dynamics of driving manoeuvrers and hazard perception, especially since hazard events unfold over several seconds, and allows systems trained on Focus100 to be deployable in such stacks. Similarly, the \qty{60}{\hertz} gaze tracking in our setup provides a sampling rate that is robust for the analysis of fixations, which are the primary correlate of a driver's perceptual information processing. Prior methodological work shows that fixation-based eye-tracking measures are accurate at \qty{60}{\hertz}, with non-significant difference in fixation detection when downsampling from high-rate data \cite{andersson2010sampling,holmqvist2011eye}. In driving research, on-road studies often use \qty{60}{\hertz} eye trackers \cite{stapel2022measuring}; with higher sampling rates mainly benefitting micro-saccade analyses \cite{leube2017sampling}. Focus100’s \qty{10}{\fps} video and \qty{60}{\hertz} gaze recording can therefore be considered well-aligned with community norms and sufficient for capturing the phenomena of interest.

\paragraph{Lab-Collected Gaze}

The ecological validity of lab-based gaze data is a common concern. Differences between passive or semi-passive viewing and active vehicle control are documented; lab protocols remove visuomotor load and can broaden scanning relative to on-road or high-fidelity simulation settings \cite{kotseruba2024data,xia2019predicting}. Controlled comparisons indicate that the magnitude of these differences is small, however, as statistical analyses in \cite{mackenzie2015eye} report modest effect sizes (though statistically significant) for changes in gaze variance when moving from video-based hazard perception to simulator driving. Critically, lab-collected gaze remains highly informative: it reliably differentiates expert from novice drivers \cite{omran2023driving}, while models trained solely on in-lab data generalise to on-road attention prediction, achieving competitive performance on real driving benchmarks \cite{xia2019predicting}. Focus100 follows this paradigm while emphasising hazardous scenarios and releasing per-subject temporal gaze streams; capturing not only where drivers look but also \emph{when}, not hitherto possible with datasets of this scale, enabling fine-grained temporal analyses of human attention and situation awareness.

\paragraph{Hazards vs.\ Crashes}

A reasonable question is whether Focus100's lack of crash events  constrain the scope of our conclusions. 
While most real-world driving is uneventful, drivers still face situations of varying risk, whereas actual crashes are rare \cite{dingus2016driver, klauer2006impact}. Crash-focused datasets are invaluable for analysing accident causation, but modelling driver attention and behaviour in the broader context of everyday driving requires coverage of non-crash yet hazardous situations.
Attentional failures, such as prolonged off-road glances or mind-wandering, often precede both crashes and near-crashes \cite{klauer2006impact, seppelt2017glass}, suggesting shared cognitive mechanisms; near-miss and sub-critical hazardous events therefore serve as effective proxies for studying driver perception and attention in safety-critical contexts \cite{kong2021patterns, seacrist2020near}.
Focus100 does not contain crashes but includes a wide range of situations with hazards of varying severity, capturing both routine and complex driving conditions where attentional demands naturally vary.
This coverage complements existing datasets, including DADA-2000 \cite{fang2021dada}, which focuses on crash prediction from in-lab attention data on crowd-sourced crash videos, and BDD-A \cite{xia2019predicting}, which uses hard braking events as hazard proxies. 
By spanning diverse hazards, Focus100 enables the study of driver attention in common critical conditions, complementing existing crash-centric datasets towards applications in automotive safety.

\section{Implementation Details}
We implemented our model using the \textit{PyTorch 2.2.1} \cite{Ansel_PyTorch_2_Faster_2024}, \textit{PyTorch Geometric 2.5.0} \cite{Fey_Lenssen_2019, Fey_etal_2025}, \textit{PyTorch Lightning 2.1.3} \cite{Falcon_PyTorch_Lightning_2019}, and \textit{ClearML 1.14.4} \cite{clearml} frameworks. Here we report some implementation specifics.

\subsection{Gaze Processing}

Our method learns using minimally preprocessed gaze data in our gaze-centric scene graphs. Here we report that process, as well as that for converting gaze sequences into fixation and saliency representations.

\paragraph{Preprocessing}
Our minimal preprocessing stage consisted of linearly interpolating across samples deemed as blinks (as detected by the Tobii tracker), ensuring the temporal continuity of the gaze signal. More specifically, gaze positions during blinks were set to \textit{(NaN, NaN)}, and the \textit{interp} function from \textit{numpy} \cite{harris2020array} was applied independently for the \textit{x} and \textit{y} coordinates to replace all invalid values. The same procedure was applied to the gaze data from the MAAD dataset. This process was carried out in the data's native gaze sampling frequency (\eg \qty{60}{\hertz} for Focus100), before downsampling by linear interpolation to align with the desired temporal scene graph frequency (as described in Sec 5.1 of the main paper).

\paragraph{Temporal sampling at 20Hz vs 10Hz}
Focus100 videos are recorded at \qty{10}{\fps}, while gaze is acquired at a higher native rate with multiple gaze samples per displayed video frame. For scene-graph construction, we represent \qty{1}{\second} windows using \qty{20}{} timesteps (\qty{20}{\hertz}). To align modalities, we upsample the video stream from \qty{10}{\fps} to \qty{20}{\fps} by frame duplication (each video frame is repeated once), and we downsample gaze to \qty{20}{\hertz} so that every timestep contains synchronised traffic-object, road-structure, and gaze nodes.

We choose \qty{20}{\hertz} to support reliable fixation-based evaluation: using a minimum fixation duration of \qty{100}{\milli\second}, low sampling rates can under-sample short fixations and distort estimated fixation statistics. We therefore ablate the effective gaze sampling rate across Focus100 and observe fixation rates of 1.70\,s$^{-1}$ at \qty{60}{\hertz}, 1.64\,s$^{-1}$ at \qty{20}{\hertz}, 1.23\,s$^{-1}$ at \qty{10}{\hertz}, and 0.59\,s$^{-1}$ at \qty{5}{\hertz}, confirming that \qty{10}{\hertz} is inadequate for fixation analysis in our setting. Importantly, the video upsampling is used only for temporal synchronisation with the \qty{20}{\hertz} graph; it does not introduce new visual content beyond the original \qty{10}{\fps} frames.

\paragraph{Postprocessing}

While our method learns from this minimally processed data, we also implement training-free post-processing to generate fixations and saliency map estimates. An identical process is also used to turn raw ground-truth human gaze sequences into saliency maps for training several baseline saliency estimation approaches.

To detect fixations in gaze sequences we apply the EyeMMV algorithm \cite{krassanakis2014eyemmv}. EyeMMV is a two-stage, dispersion-based fixation detector (I-DT). Subsets of samples are preliminarily classified as fixations where spatial dispersion remains below a coarse threshold; when this bound is exceeded, the segment is refined with a stricter dispersion threshold to trim edge samples. The candidate is then accepted as a fixation if its duration surpasses a minimum, with inter-fixation intervals labelled as saccades and fixation position defined by the centroid of accepted samples. In our setup we use thresholds \(t_0=0.08\) and \(t_1=0.05\) (in normalised image space), and enforce a minimum fixation duration of \qty{0.1}{\second}; detected fixations were additionally manually spot-checked on a subset of trials.

Saliency maps are generated by first accumulating fixations onto a 2D grid matching the spatial resolution of the input frame, where each pixel value represents the number of fixation samples falling at that location (after rounding coordinates to the nearest integer), aggregated across all subjects or generated sequences corresponding to that frame. The resulting discrete fixation map is then smoothed with a Gaussian filter using a standard deviation of $\sigma = 19 \times (w / 640)$, where $w$ is the frame width, following \cite{djilali2024learning}. Finally, the saliency map is normalised by its maximum value, yielding intensity values in the range $[0, 1]$.

\begin{figure*}[t!]
    \centering
    \begin{subfigure}{\textwidth}
        \centering
        \includegraphics[width=\linewidth]{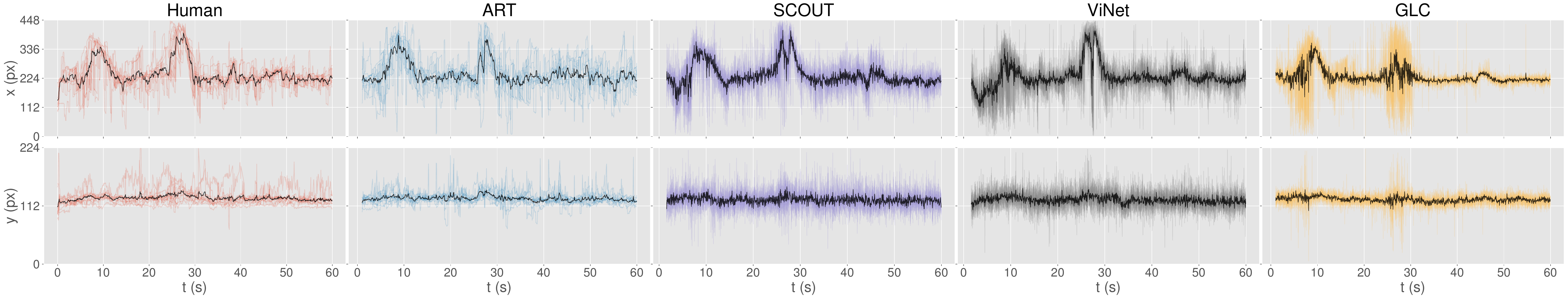}
        \label{fig:subfig1}
    \end{subfigure}
    \hfill
    \begin{subfigure}{\textwidth}
        \centering
        \includegraphics[width=\linewidth]{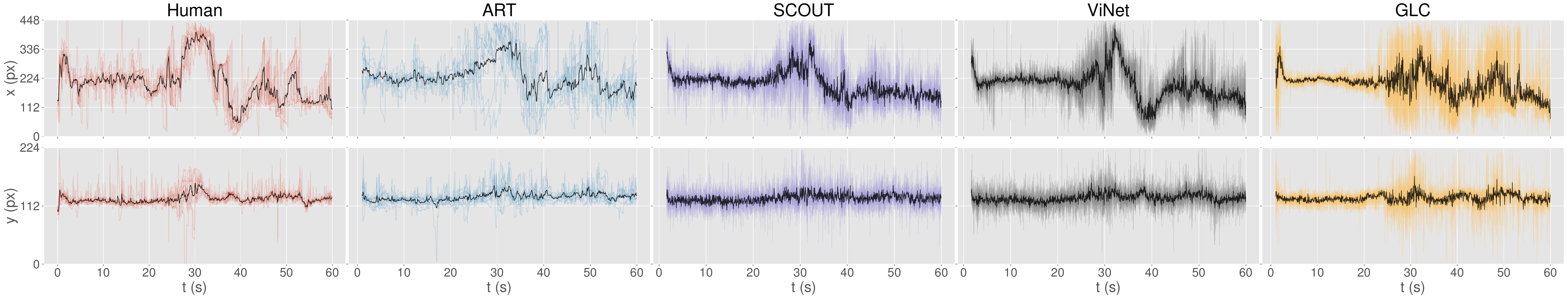}
        \label{fig:subfig2}
    \end{subfigure}
    \hfill
    \begin{subfigure}{\textwidth}
        \centering
        \includegraphics[width=\linewidth]{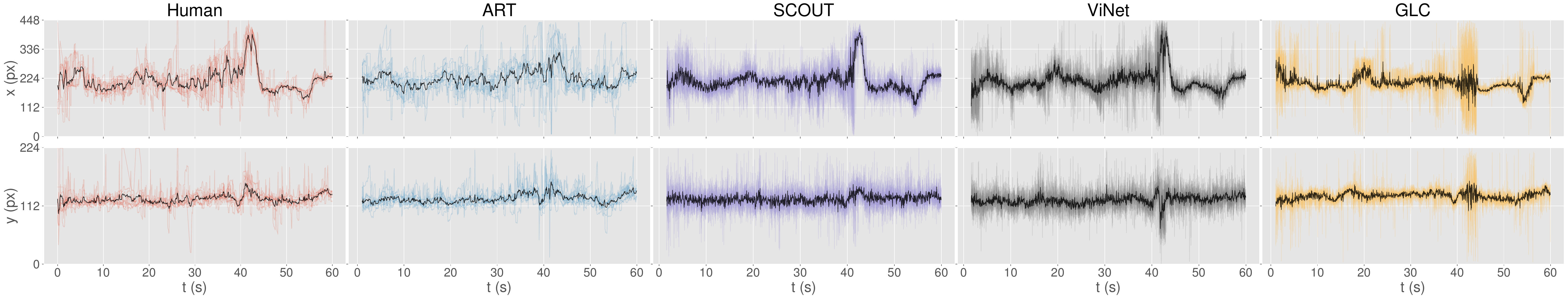}

        \label{fig:subfig3}
    \end{subfigure}
    \caption{Human gaze sequences compared to generated sequences by ART, SCOUT, ViNet and GLC on 3 videos from the test set. We plot the $x$ and $y$ positions of gaze over time separately, including the $y$-axis for completeness as it was not shown in the main text. Each line represents a sampled gaze sequence, with the mean gaze sequence shown in black.}
    \label{fig:xy_gaze_plots}
\end{figure*}

\subsection{Scene Graph Construction}
As mentioned in Sec 5.1 of the paper, we used the \textit{YOLOv8x} \cite{Jocher_Ultralytics_YOLO_2023} detector to obtain the object bounding boxes. The appearance features were extracted from the 12th layer of a pretrained \textit{vgg16\_bn} network \cite{simonyan2014very} using \textit{ROIAlign} \cite{he2017mask}, yielding a 128-D appearance vector. The \textit{structure} node is obtained by estimating the drivable-area mask with \textit{YOLOPv2} \cite{han2022yolopv2}, resizing the mask to \numproduct{16 x 8}~\si{\px}, and flattening. Each object's depth was estimated using \textit{monodepth2} \cite{godard2019digging} as the mean of inverse disparity within the object's bounding box.

\textbf{Input node vectors:} The dimensionality of each input node vector used in our experiments is 144: the object's $x$ and $y$ coordinates (2), its bounding box shape (2), the detector detection score (1), the appearance vector (128), depth estimate (1), and the label one-hot encoding (10; `car', `person', `bicycle', `motorcycle', `bus', `truck', `traffic light', `stop sign', `gaze node', `structure node').

\textbf{Input edge vectors:} 
The dimensionality of input edge vectors used in the experiments is 5: 3D positional difference between the connecting nodes (3; $x$, $y$, depth), timestep difference (1), and cosine similiarity between the node appearance vectors (1).

\textbf{Temporal connectivity:} Nodes are connected temporally if the timestep difference between the nodes is included in the set $\mathcal{T}_d=\{1,2,4,8,16\}$.

\subsection{Graph Processor}
The Graph Processor processes the input scene graph as described in Sec 3.2 of the paper. Here we provide additional details.

\textbf{Node embeddings:} The dimensionality used for the node-type-specific linear embeddings of the node vectors is $d=128$. Each node's timestep is encoded as alternating sine and cosine waves and added to the embedding.

\begin{figure}[t]
    \centering
    \includegraphics[width=\columnwidth]{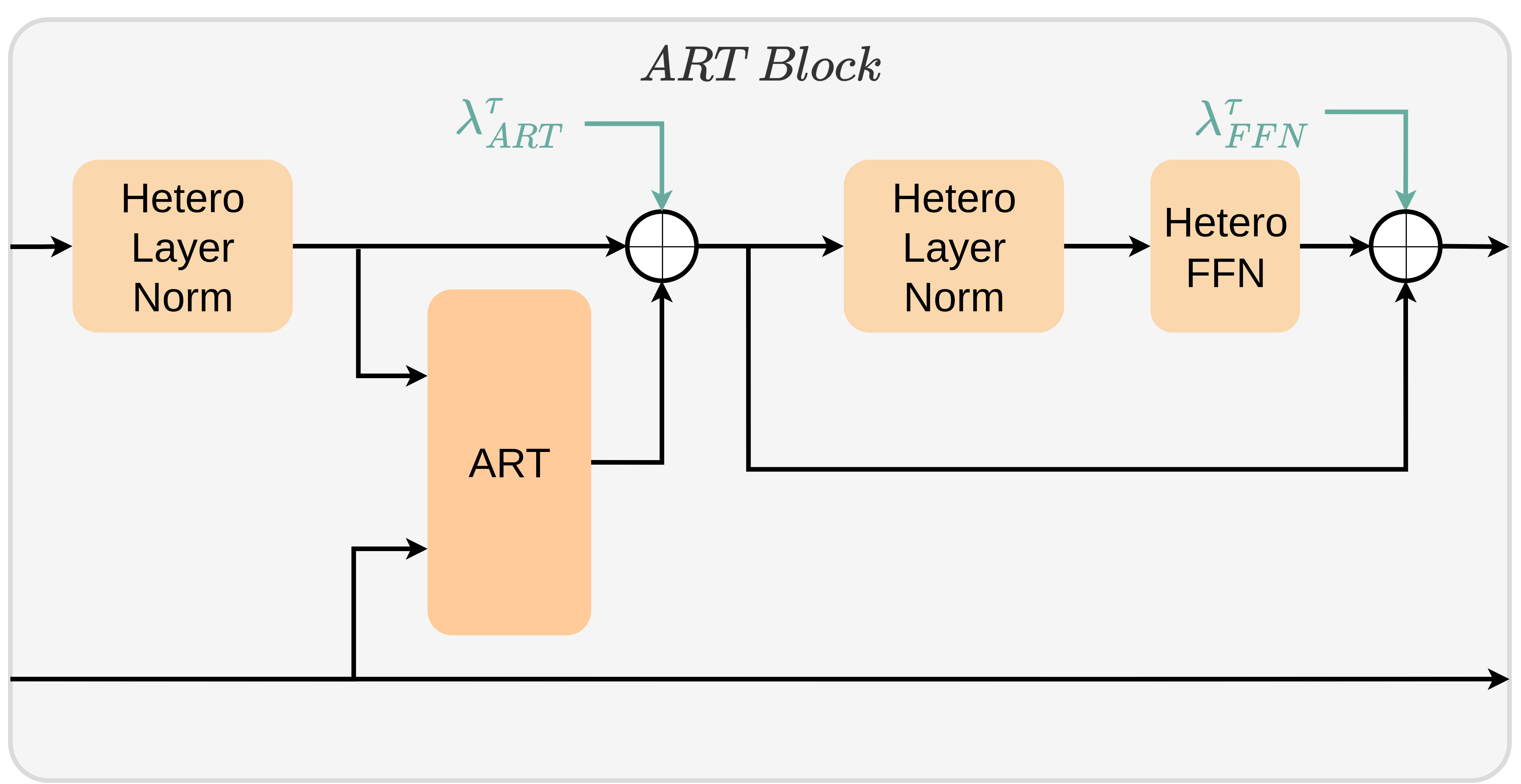}
    \caption{Illustration of the ART block. The block applies LayerNorm, ART attention with a residual connection, another LayerNorm, followed by a two-layer feed-forward network (FFN) with a second residual connection. $\lambda^\tau_{\text{ART}}$ and $\lambda^\tau_{\text{FFN}}$ denote the node-type-specific learnable parameters controlling the strengths of the residual connections, $0\leq\lambda^\tau_{\text{ART}},\lambda^\tau_{\text{FFN}}\leq1$.}
    \label{fig:artblock}
\end{figure}

\textbf{ART block:}
We use $L=2$ ART blocks in the Graph Processor in our experiments. An illustration of an ART block is shown in \Cref{fig:artblock}.
The edge vectors in ART (relative affinity $\mathbf{a}_{i, j}$ in Fig.\ 3 in the main paper) are embedded into key and value embeddings 
($\mathbf{p}_{i,j}^{K}$ and $\mathbf{p}_{i,j}^{V}$ in Eqs.\ (6) and (7)) 
using two independent MLPs, each implemented as a node-type dependant linear layer followed by \textit{BatchNorm}, a \textit{ReLU}, and another node-type dependant linear layer. Both linear projections output $d$-dimensional vectors, with $d=128$.
The query, key and value vectors, $\mathbf{Q}_i$, $\mathbf{K}_j$, and $\mathbf{V}_j$, are calculated using a node-type-specific linear layer with a bias, outputting a \numproduct{3 x 128}-dimensional vector which is then split into three 128-dimensional vectors.

\textbf{FFN:} The feed-forward network is implemented as two node-type-specific linear layers with biases; the first outputs a 256-dimensional vector, which is passed through a ReLU, and the second linear layer outputs a 128-dimensional vector. A residual connection with a node-type-specific learnable parameter $\lambda^\tau_{\text{FFN}}$ is used. See the ART Block illustration in \Cref{fig:artblock}.

\subsection{Object Density Network}

The updated node features belonging to the last timestep in the input spatio-temporal scene graph are fed into the ODN to estimate the parameters of a GMM modelling the future gaze position probability distribution. The parameters predicted by the ODN are listed in Sec 3.3 of the paper and are the output of a node-type-specific linear layer.

\section{Qualitative Results}

\paragraph{Gaze Sequences} The visualisations of sampled gaze in the main paper only show the horizontal position of gaze plotted against time. In \cref{fig:xy_gaze_plots} we include a number of plots in which both the $x$ and $y$ dimensions of sampled gaze over time are shown.

\paragraph{2D Sequences}

\begin{figure*}[htbp]
    \centering
    \includegraphics[width=\linewidth]{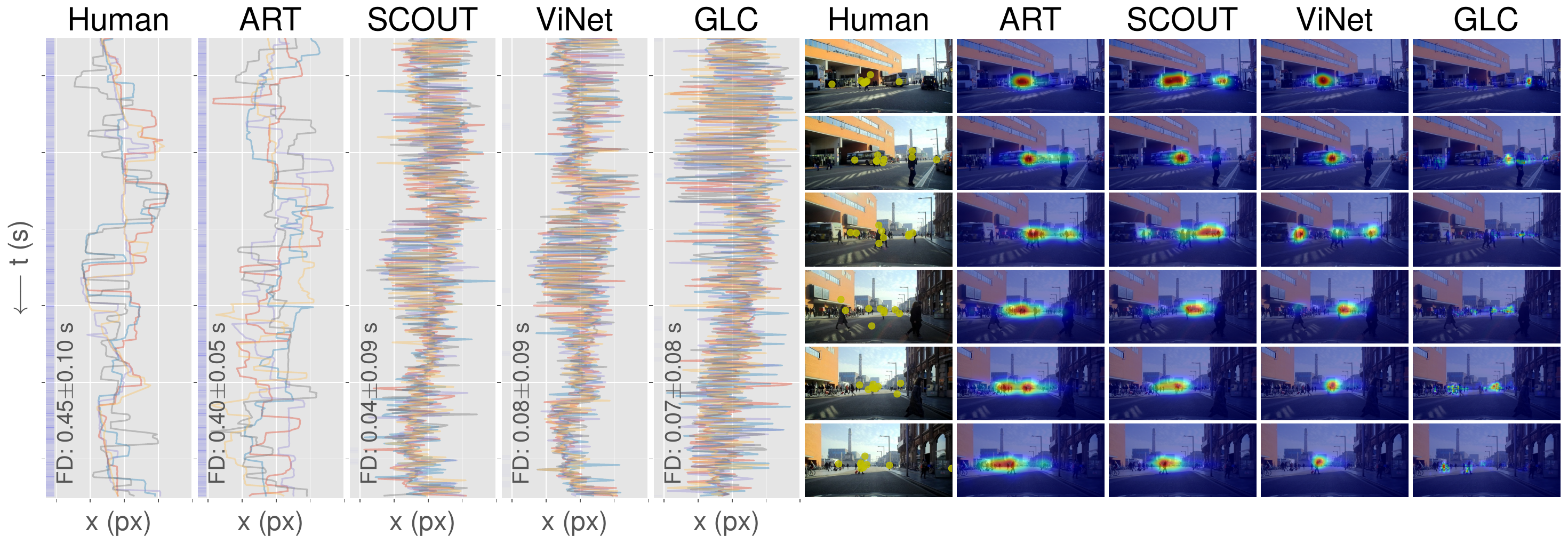}
    \caption{Preceding frames from the same test subsequence  show that the largest values in the ART saliency maps are concentrated at positions corresponding to the ground truth gaze points. 
    As shown in the left part of the figure, the gaze dynamics generated by ART closely follow human gaze patterns, whereas the SCOUT, ViNet, and GLC sequences exhibit notably more volatile behaviour.}
    \label{fig:good1}
\end{figure*}

\begin{figure*}
    \centering
    \includegraphics[width=\linewidth]{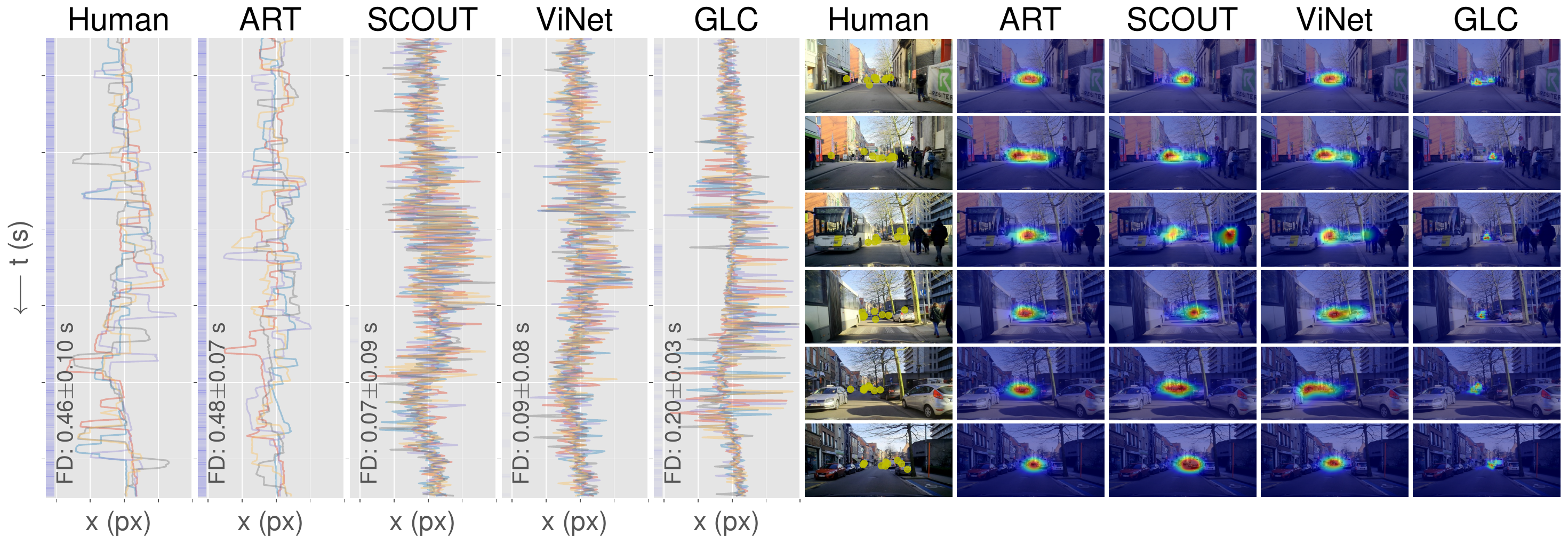}
    \caption{Another example sequence from the unseen test set is presented. Consistent with the previous examples, the samples generated by ART closely follow the ground-truth human gaze dynamics, while the other methods exhibit unsteady and less realistic gaze behaviour throughout the sequence. 
    The saliency maps produced by ART closely reflect the distribution of ground-truth human gaze points. In contrast, GLC produces low-variance saliency maps concentrated near the centre of the road. SCOUT and ViNet generate saliency maps that are qualitatively similar to those of ART, except in the third row, where they highlight the pedestrians on the pavement on the right.}
    \label{fig:good2}
\end{figure*}

\begin{figure*}[htbp]
    \centering
     \includegraphics[width=\linewidth]{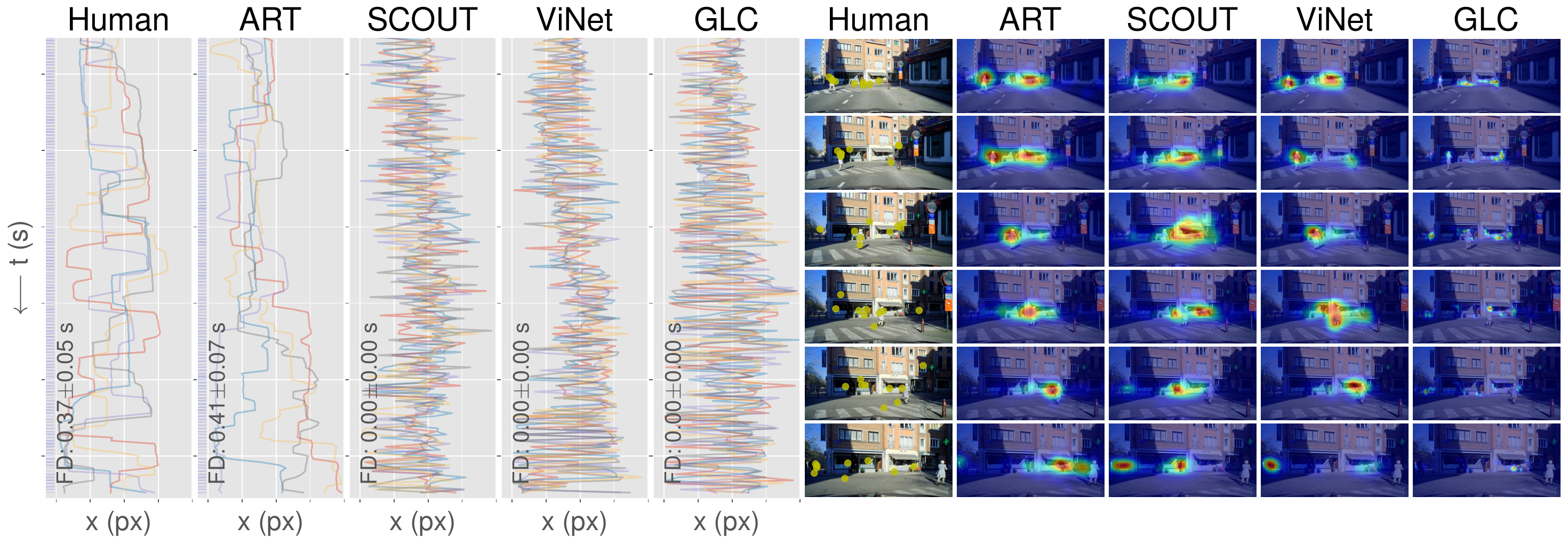}
    \caption{An example of a failure case for our method. The ART gaze sequence samples shown on the left indicate that the simulated gaze tends to follow the pedestrian crossing the road, in contrast to the ground truth human gaze sequences.}
    \label{fig:pedestrian}
\end{figure*}

\cref{fig:good1,,fig:good2} display more sampled gaze sequences and saliency maps generated by ART and the baseline models on additional unseen test videos, along with human gaze sequences for comparison. We consistently show that gaze sequences generated by ART closely mimic the human gaze behaviour. We show a failure case of our method in \cref{fig:pedestrian}.

\begin{figure*}[htb]
    \centering
    \includegraphics[width=\linewidth]{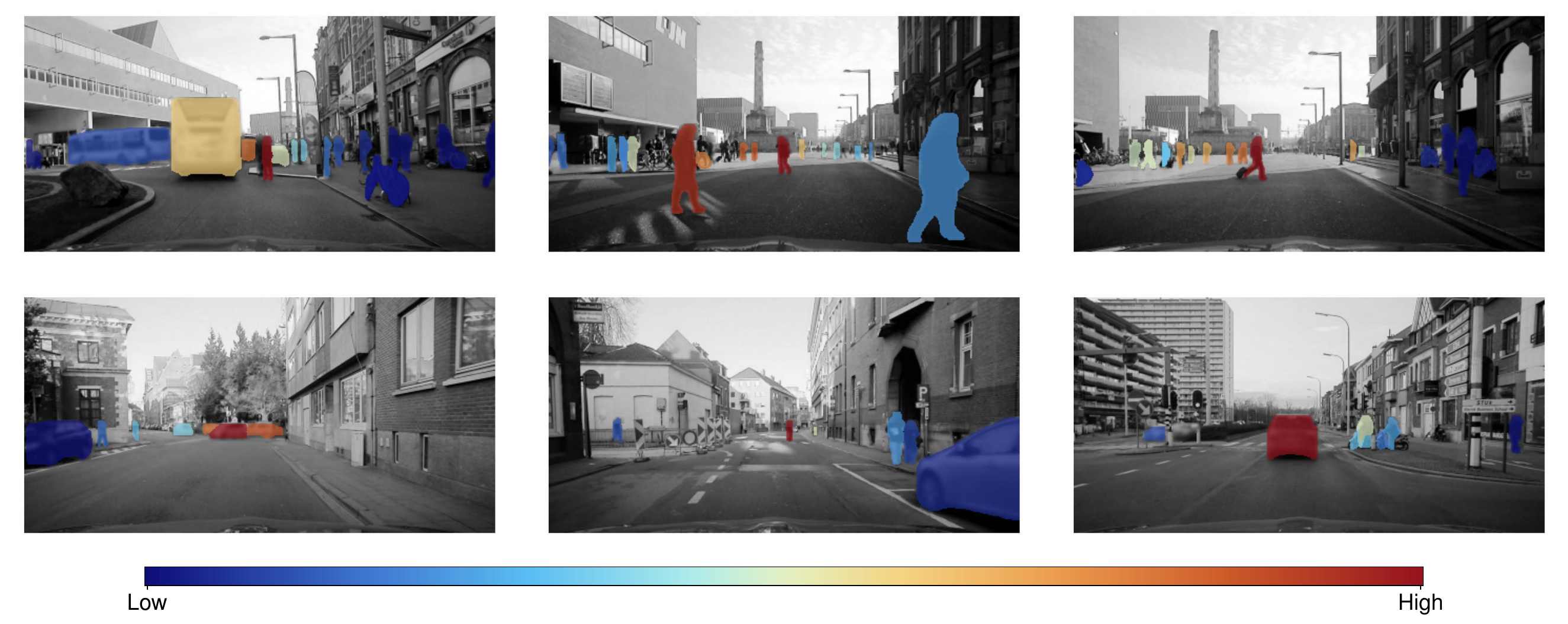}
    \caption{Object salience ranking estimated as the average mixing node weight calculated over 60 ART simulation runs. The node mixing weights are further normalised by the maximum mixing weight in the given frame. Low saliency objects are shown in blue, and the most salient object is shown in red. Note that the colormaps represent rank order and are not consistent across images. See under \textit{Object Salience} in \Cref{ref:object_salience} for details.}
    \label{fig:saliency}
\end{figure*}

\paragraph{Object Salience}
\label{ref:object_salience}
Here we explore whether our model produces reasonable estimates of the saliency of objects within images, a task considered in \cite{deng2024advancing} for example. To estimate the saliency ranking of objects in a given frame, we perform 60 gaze sequence simulation runs using our proposed method for a specified video sequence. For each frame, we store the mixing weights estimated by the ODN for each graph node (\ie object detection). We ignore the \textit{structure} and \textit{gaze} nodes as we are only interested in the saliency of individual objects. The average mixing weight for each node in a frame is estimated by summing the mixing weights across all runs for each node, and renormalising them using softmax to account for the removed nodes. As we are interested in ranking the objects within a specific frame, we further divide the mixing weights of nodes in the frame by the maximum mixing weight in that frame. We use YOLOv8x-seg \cite{Jocher_Ultralytics_YOLO_2023} to estimate the segmentation masks for all objects contained as nodes in the graph for a given frame. We overlay the segmentation masks over the input image, assigning them a colour based on the estimated normalised mixing weight. Objects of low saliency rank within an image are shown in blue, and the most salient object(s) is highlighted in red. Example saliency rankings can be seen in \Cref{fig:saliency}.

\section{Gaze State Dynamics}

In this section we analyse the dynamics of gaze state transitions between \textit{saccades} and \textit{fixations}. We identify all timesteps marking the onset of a fixation and observe a time window of \qty{0.5}{\second} before and after this point. Each timestep within this window is labelled with a 1 if the gaze at that timestep corresponds to a fixation, or 0 otherwise (\ie, if it was a part of a saccade). We calculate the differences between consecutive elements of the described array, \ie $\mathbf{d}[t] = \mathbf{v}[t + 1] - \mathbf{v}[t]$, where we use $\mathbf{v}$ to denote the initial vector of fixations and saccades, $\mathbf{d}$ to mark the vector of differences, and $t$ to index the elements. Each value of the resulting vector will be either -1, 1 or 0, where -1 denotes a change from a fixation to a saccade, 1 marks a change from a saccade to a fixation, and 0 means no state change. Calculating the mean value of all the  vectors $\mathbf{d}_i$, constructed for each fixation in the test set, will give us an empirical expected value of the state change direction for each timestep in the observed window centered around the start of a fixation, $\mathbb{E}(\mathbf{d})$. A more positive value means a higher probability of a saccade-to-fixation state change, and a more negative value means a higher probability of a fixation-to-saccade state change. 

\begin{figure*}
\centering
    \begin{subfigure}{0.3\textwidth}
        \centering
        \includegraphics[width=\textwidth]{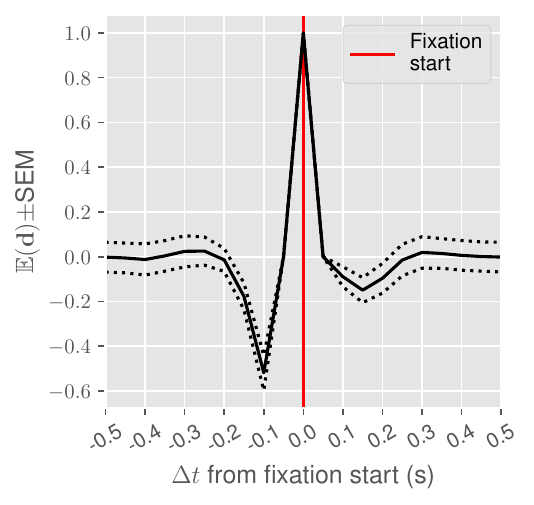}
        \caption{Gaze state dynamics estimated from ground truth human gaze samples.}
        \label{fig:human_gaze_prob}
    \end{subfigure}
    \hfill
    \begin{subfigure}{0.3\textwidth}
        \centering
        \includegraphics[width=\textwidth]{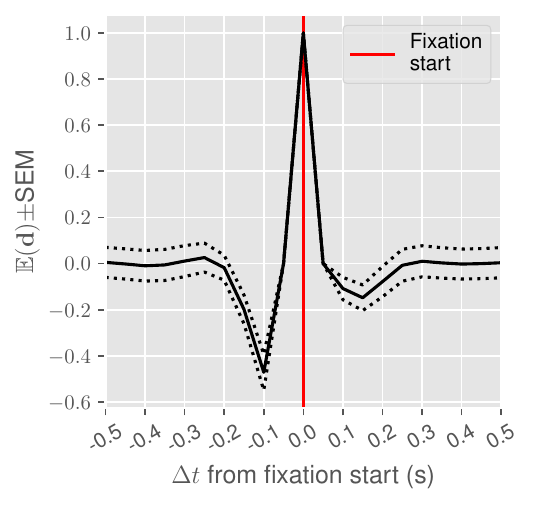}
        \caption{Gaze state dynamics estimated from ART samples.}
        \label{fig:art_gaze_prob}
    \end{subfigure}
    \hfill
    \begin{subfigure}{0.3\textwidth}
        \centering
        \includegraphics[width=\textwidth]{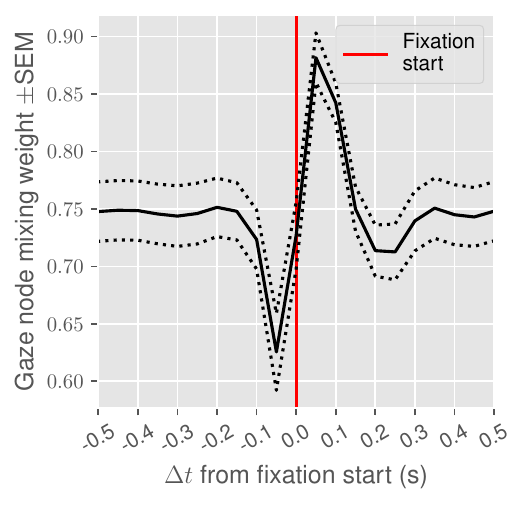}
        \caption{ART/ODN gaze node mixing weight as fixation mechanism.}
        \label{fig:model_meachanism}
    \end{subfigure}
    \caption{Evolution of the expected value of the gaze state change direction as a function of time relative to the start of a fixation. In \cref{fig:human_gaze_prob,,fig:art_gaze_prob} positive values on the y axis denote a higher probability of the gaze state changing from a \textit{saccade} to a \textit{fixation}, while negative values indicate a higher probability of the reverse. The start of a fixation (at $\Delta t=0$) is preceded by an increased probability of a saccade (the dip on the left), and followed by another slight increase of the saccade probability. The gaze state change probability dynamics are very similar when estimated for the ground truth \textbf{human} gaze samples (\cref{fig:human_gaze_prob}) and using the samples from \textbf{ART} (\cref{fig:art_gaze_prob}).
    In \cref{fig:model_meachanism} we show the \textbf{model mechanism}, \ie gaze node mixing weight as a function of time relative to a fixation start. High gaze node mixing weight means a high probability of the gaze maintaining the same location in the next timestep, while a lower mixing weight means a higher probability of the gaze changing its location in the next timestep. Note the similarity of this plot to the ones shown in \cref{fig:human_gaze_prob,,fig:art_gaze_prob}. The shift on the $x$-axis towards the right is due to the gaze node mixing weight affecting the gaze state at the \textit{next timestep}.} 
    \label{fig:gaze_state_dynamics}
\end{figure*}

In \Cref{fig:human_gaze_prob,,fig:art_gaze_prob} we plot this expected value $\mathbb{E}(\mathbf{d})$ as a function of time centered at the beginning of a fixation, estimated using ground truth human gaze samples and samples generated by ART, respectively. We can see that the plots for both the human gaze and our method closely resemble each other. The initial dip preceding the start of a fixation denotes an increase of probability of a state change from a fixation into a saccade; first a saccade needs to happen for a fixation to start, \ie the probability of a saccade needs to increase. The probability of a saccade-to-fixation state change is the highest when the fixation is actually starting, shown at $\Delta t=0$. This is then followed by another drop, denoting a slightly increased probability of another saccade occurring.

\paragraph{ART/ODN Fixation Mechanism}
In \Cref{fig:model_meachanism} we plot the gaze node mixing weight as a function of time since the start of a fixation. A high gaze node mixing weight implies a higher probability that the gaze in the next timestep stays at the same location (a part of a fixation), while a lower gaze node mixing weight means an increased probability of a saccade occurring at the next timestep. Notice the resemblance of the plotted shape and the shape of the signal in \cref{fig:human_gaze_prob,,fig:art_gaze_prob}, suggesting the gaze node mechanism for producing fixations worked as intended. As the gaze node mixing weight affects the gaze state in the \textit{next timestep}, the signal appears to be shifted one step to the right compared to the gaze state change dynamics plots in \cref{fig:human_gaze_prob,,fig:art_gaze_prob}.

\section{Spectral Analysis of Gaze Variance}

To assess whether models reproduce the temporal structure of human inter-observer variability, we analyse the power spectral density (PSD) of residual gaze trajectories relative to each group’s own mean trajectory. For each test sequence $s$ and group $g \in \{\mathrm{Human}, \mathrm{ART}, \mathrm{SCOUT}, \mathrm{VINET}, \mathrm{GLC}\}$, we compute the group mean trajectory
\begin{equation}
\bar{\mathbf{p}}_{g}^{(s)}(t) = \frac{1}{N_g^{(s)}} \sum_{i=1}^{N_g^{(s)}} \mathbf{p}_{g,i}^{(s)}(t),
\end{equation}
and define residual trajectories
\begin{equation}
\mathbf{r}_{g,i}^{(s)}(t) = \mathbf{p}_{g,i}^{(s)}(t) - \bar{\mathbf{p}}_{g}^{(s)}(t).
\end{equation}
We compute the scalar residual magnitude $r(t) = \|\mathbf{r}(t)\|$ and estimate its PSD using Welch’s method. The integral of the PSD corresponds to total within-group residual variance, while its distribution over frequency reflects the temporal organisation of that variance.

\begin{figure}[t]
    \centering
    \includegraphics[width=\columnwidth]{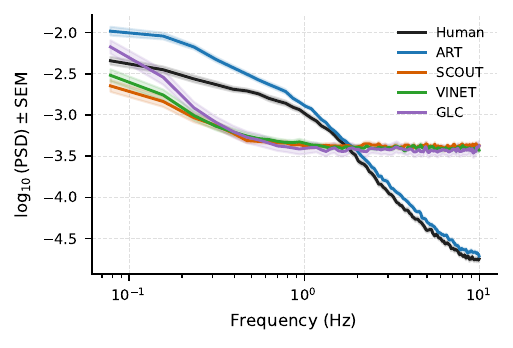}
    \caption{Residual PSD across all shared test sequences. Curves show mean \(\log_{10}\) PSD with \(\pm\)SEM across all sequences for Human, ART, SCOUT, VINET, and GLC. ART most closely tracks the human spectral variance profile, consistent with the dynamics results presented in the main text.}
    \label{fig:residual_psd_all_sequences}
\end{figure}

Figure \ref{fig:residual_psd_all_sequences} shows the residual PSD averaged across test sequences. Human residuals exhibit a low-frequency-dominated spectrum, indicating that inter-observer variability is dominated by lower-frequency components rather than short-timescale fluctuations. ART closely matches the human spectral profile across frequencies. In contrast, SCOUT, VINET, and GLC show comparatively reduced low-frequency power and flatter spectra, indicating less temporally structured variability.

To summarise spectral structure per sequence, we compute the ratio
\begin{equation}
r_{g,i}^{(s)} =
\frac{\int_{1}^{5} \mathrm{PSD}_{g,i}^{(s)}(f)\,df}
     {\int_{0.1}^{1} \mathrm{PSD}_{g,i}^{(s)}(f)\,df},
\end{equation}
where the lower band (0.1--1 Hz) captures slower residual dynamics and the higher band (1--5 Hz) captures faster fluctuations. The ratio therefore reflects the relative contribution of fast versus slow components of within-group variability. Sequence-level group statistics are obtained by averaging across samples.

Across the 20 test sequences, ART consistently exhibited the smallest deviation from the human spectral ratio. We performed paired one-sided Wilcoxon signed-rank tests comparing ART against each alternative model under the hypothesis $d_{\mathrm{ART}} < d_{\mathrm{other}}$, with Holm correction for multiple comparisons. All comparisons were significant after correction ($p < 0.001$), with complete directional consistency across sequences.

These results indicate that human inter-observer variability is temporally structured and dominated by slower components. Among evaluated models, ART closely reproduces the spectral organisation of this variance across simulated gaze trajectories on the same sequence, whereas alternative models exhibit significantly different spectral signatures.

\section{Latency-Accuracy Trade-Off}

Simulations using the method presented in the paper run at an average of \qty{68}{\ms} per frame (\qty{15}{\fps}), using input data of resolution of \numproduct{448 x 224}~\si{\px}, on a single L40S GPU, including the perception stack, graph construction, ART and ODN. Profiling shows runtime split between perception (43\%) and ART+ODN (57\%). 

\begin{figure}[t]
    \centering
    \includegraphics[width=\columnwidth]{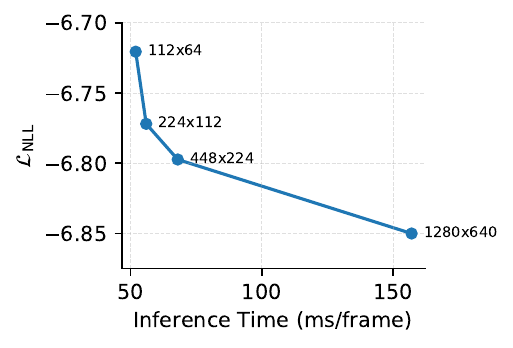}
    \caption{Negative log-likelihood loss across the test set against the time taken to process each frame and generate a next-step gaze position. The used input video resolution is shown next to each data point.}
    \label{fig:latency_accuracy_tradeoff}
\end{figure}

We conducted a latency-accuracy trade-off analysis by varying the input resolution of the method, results of which can be seen in Figure \ref{fig:latency_accuracy_tradeoff}. The curve shows a consistent accuracy-latency trade-off: increasing input resolution improves predicted gaze likelihood (lower $\mathcal{L}_{\mathrm{NLL}}$) but the gains taper off as resolution increases. Most of the improvement is achieved when moving from very low to mid resolutions, while further increasing resolution produces relatively smaller additional accuracy benefit. The \numproduct{448 x 224}~\si{\px} setting is a good mid-point for real-time analysis, whereas full resolution \numproduct{1280 x 640}~\si{\px}, running at \qty{6}{fps}, is better suited to offline use when latency is less critical.

\section{Baseline Details}

We used publicly available official implementations\footnote{
GLC: \url{https://github.com/BolinLai/GLC/}, SCOUT: \url{https://github.com/ykotseruba/SCOUT}, ViNet: \url{https://github.com/samyak0210/ViNet},
DReyeVENet: \url{https://github.com/ndrplz/dreyeve}. 
} of GLC \cite{lai2024eye}, SCOUT \cite{kotseruba2024understanding}, ViNet \cite{jain2021vinet} and DReyeVENet \cite{palazzi2018predicting} in our experiments.
We give the training and inference details for these methods below.

\subsection{Global-Local Correlation (GLC)}

\paragraph{Training}

The \textit{GLC} model \cite{lai2024eye} is the current state-of-the-art model in egocentric gaze estimation. It is trained on sequences of 8 temporally equidistant square (\numproduct{256 x 256}~\si{\px}) crops from the input RGB video, which is resized to a height of \qty{256}{\px} while maintaining the aspect ratio. We follow the training procedure from \cite{lai2024eye} with the \textit{MViT} \cite{fan2021multiscale} architecture as the backbone network, initialised with weights pretrained on the \textit{Kinetics-400} \cite{kay2017kinetics} dataset. 
We use a temporal sampling rate of 3 in our experiments, \ie we sample 8 frames from a 22-frame window with equal spacing, and take the last frame's predicted gaze map as the model output.
We train the model for 25 epochs, with a base learning rate set to $\SI{5e-5}{}$.
We use batch size $16$ and run the training on 2 \textit{NVIDIA GeForce RTX 3090} GPUs.

\paragraph{Inference}
At inference, the original approach only produces gaze probability maps for the central crop of the input; to create our rectangular maps we slide the \numproduct{256 x 256}~\si{\px} cropping region horizontally over the rectangular input with a \qty{16}{\px} stride and average the results. To produce sequences using GLC we sample the output gaze probability map for each frame of our sequences.

\subsection{SCOUT}

\paragraph{Training}
We train the SCOUT model using the task-free configuration, with an input clip length of 16 frames and image resolution of \numproduct{224 x 224}~\si{\px}. The encoder consists of 4 layers with a pretrained and trainable Video Swim Transformer \cite{liu2022video} backbone. Training is performed for up to 10 epochs using the Adam optimiser with a learning rate of $\SI{1e-4}{}$, a batch size of 4, and early stopping enabled. Learning rate scheduling is applied, and the model achieving the lowest validation loss is used for inference.

\paragraph{Inference}

Inference also follows the official SCOUT implementation. Given an input sequence of 16 frames from the test set, the predicted saliency map produced by the trained SCOUT model is reshaped to match the size of the ground truth saliency map (\numproduct{448 x 224}~\si{\px}), it is blurred using a Gaussian kernel of size \numproduct{11 x 11}~\si{\px}. We normalise the predicted saliency map by dividing it by its maximum value. This saliency map is used as the prediction for the last frame in the input sequence.

\subsection{ViNet}

\paragraph{Training}

We train ViNet without the audio modality with the clip size fixed to 16 frames. The optimiser is Adam with learning rate $\SI{1e-4}{}$, batch size 8, and training for up to 40 epochs. Learning-rate scheduling was disabled. The architecture uses the S3D network \cite{xie2018rethinking} as the video encoder, pre-trained on the \textit{Kinetics-400} \cite{kay2017kinetics} action-recognition dataset. The best model is selected as the one with the lowest validation loss.

\paragraph{Inference}

We evaluate ViNet the same way as SCOUT; given an input clip of 16 frames, the model predicts a saliency map corresponding to the last frame. This predicted map is blurred by a \numproduct{11 x 11}~\si{\px} Gaussian kernel and normalised by dividing it by its maximum value. To obtain saliency predictions across the entire sequence, a sliding window approach is used, generating overlapping 16-frame input clips. This is repeated on the whole test set.

\subsection{DReyeVENet}

\paragraph{Training}

We used the official implementation from the DReyeVENet repository in our experiments. Only the image saliency branch was trained, while the optical flow and semantic segmentation branches were excluded, as the pre-trained segmentation model weights were not publicly available. Focusing on the image branch enabled efficient experimentation and ensured a stable, reproducible setup while maintaining a representative subset of the original architecture. The batch size used was set to 4, `train samples per epoch' was set to 8192, and the learning rate was set to $\SI{5e-5}{}$. Clips of 16 frames were used as input, normalised by subtracting the mean frame value estimated from the training set. All the frames were resized to \numproduct{448 x 448}~\si{\px} to match the original DReyeVENet training setup.

\paragraph{Inference}

The model with the lowest validation loss was used for evaluation. We follow the official testing code to generate saliency maps for entire sequences in the test set. As with the other baseline methods, a sliding-window approach with a clip size of 16 frames was employed to produce predictions for all frames within each sequence.

\section{MAAD Dataset Splits}

The MAAD dataset \cite{gopinath2021maad}
defines only a training and testing split (80\% / 20\%). In our experiments, we further divide the training set into training and validation subsets (87.5\% / 12.5\%), ensuring no overlap between any of the splits.

\newpage

{
    \small

    \section*{Acknowledgements}
        \noindent This work was funded by Toyota Motor Europe. We thank Catriona Rutter for her assistance with the collection and annotation of the Focus100 dataset.  
        
    \bibliographystyle{ieeenat_fullname}
    \bibliography{references}
}

\end{document}